\documentclass[10pt, a4paper]{article}

\usepackage{booktabs}
\usepackage{makecell}
\usepackage{multirow}
\usepackage{subcaption}
\usepackage{fontawesome5}

\newcommand{\lmu}{\faMountain}
\newcommand{\mcml}{\faRobot}

\newcommand{\inqaplus}{I\textsc{n}QA+}
\newcommand{\geniqa}{G\textsc{en}\-IQA}
\newcommand{\remapped}{\textsc{re\-mapped}}
\newcommand{\yesno}{\textsc{yes\-no}}

\newcommand{\labelone}{\textit{Yes}}
\newcommand{\labeltwo}{\textit{No}}
\newcommand{\labelthree}{\textit{Conditional Yes}}
\newcommand{\labelfour}{\textit{Neither Yes nor No}}
\newcommand{\labelfive}{\textit{Other}}
\newcommand{\labelsix}{\textit{Lacking Context}}
\newcommand{\flagde}{\includegraphics[width=0.5cm]{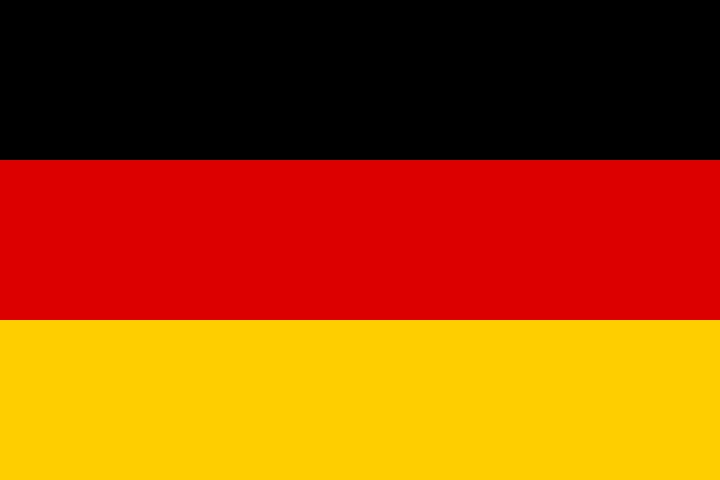}}
\newcommand{\flaguk}{\includegraphics[width=0.5cm]{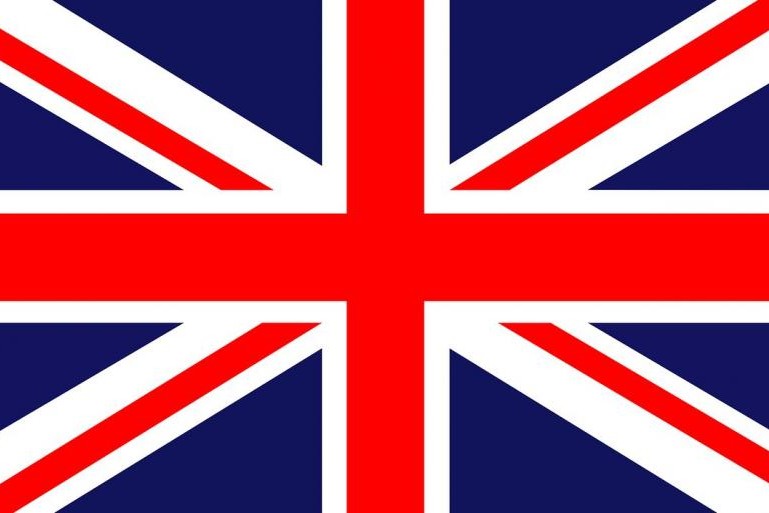}}
\newcommand{\flagba}{\includegraphics[width=0.5cm]{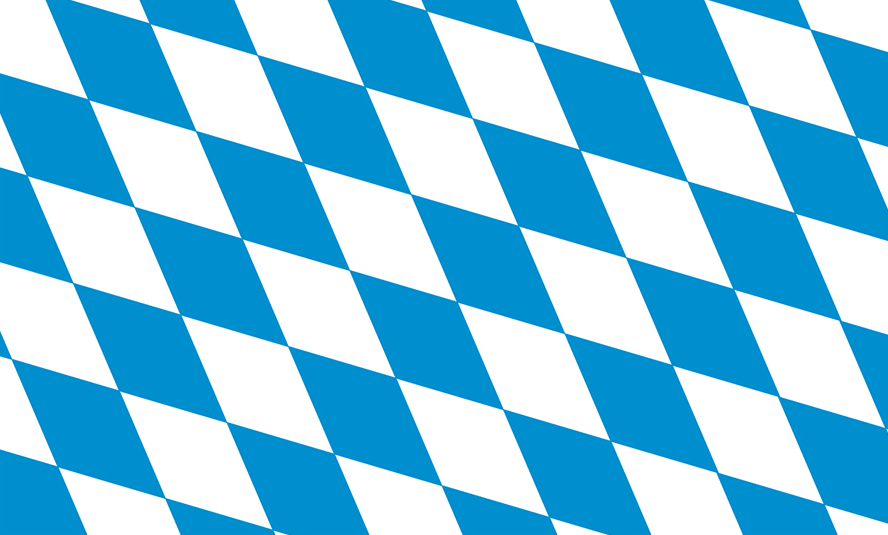}}
\newcommand{\flagmulti}{\includegraphics[width=0.5cm]{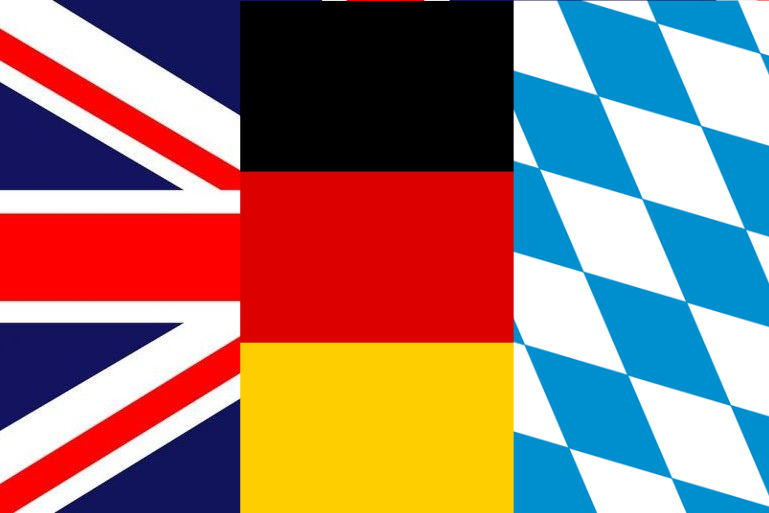}}
\newcommand{\stdev}[1]{\small{\textcolor{gray}{$\pm$#1}}}
\newcommand{\checksmall}{\includegraphics[width=0.3cm]{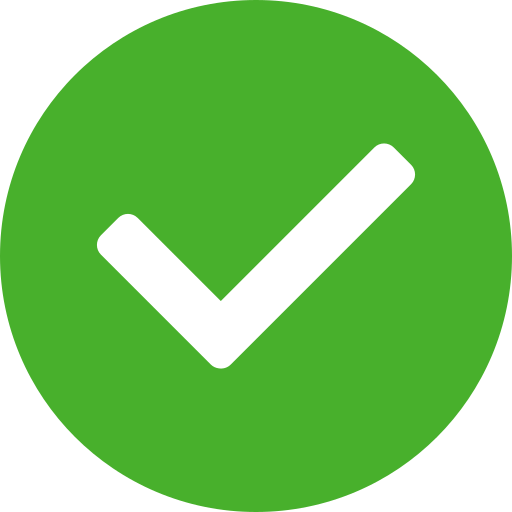}}
\newcommand{\checkyellow}{\includegraphics[width=0.3cm]{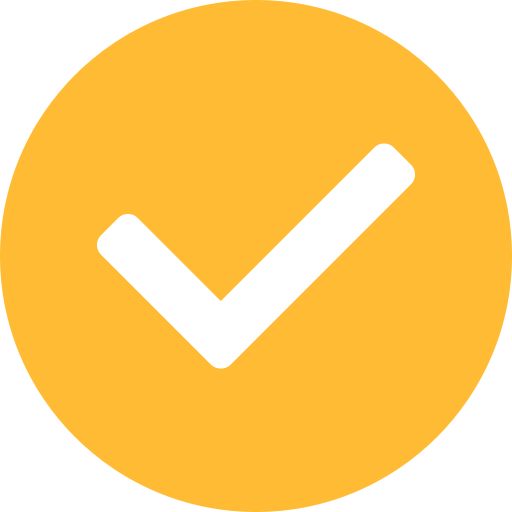}}
\newcommand{\crosssmall}{\includegraphics[width=0.3cm]{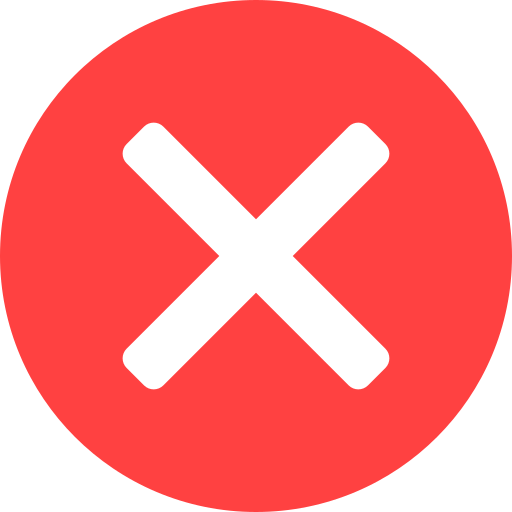}}

\newcommand{\repo}{\href{https://github.com/mainlp/Multilingual-IQA}{https://github.com/mainlp/Multilingual-IQA}}

\usepackage[final]{lrec2026} %

\title{Indirect Question Answering in English, German and Bavarian: \\A Challenging Task for High- and Low-Resource Languages Alike}

\name{Miriam Winkler\textsuperscript{\lmu}, Verena Blaschke\textsuperscript{\lmu\mcml}, Barbara Plank\textsuperscript{\lmu\mcml}} 

\address{\textsuperscript{\lmu}MaiNLP lab, Center for Information and Language Processing, LMU Munich, Germany \\
\textsuperscript{\mcml}Munich Center for Machine Learning, Germany\\
         mi.winkler.wald@gmail.com, verena.blaschke@cis.lmu.de, b.plank@lmu.de\\
         }

\abstract{
Indirectness is a common feature of daily communication, yet is underexplored in NLP research for both low-resource as well as high-resource languages. Indirect Question Answering (IQA) aims at classifying the polarity of indirect answers. 
In this paper, we present two multilingual corpora for IQA of varying quality that both cover English, Standard German and Bavarian, a German dialect without standard orthography: \inqaplus, a small high-quality evaluation dataset with hand-annotated labels, and
\geniqa, a larger training dataset, that contains artificial data generated by GPT-4o-mini.
We find that IQA is a pragmatically hard task that comes with various challenges, based on several experiment variations with multilingual transformer models (mBERT, XLM-R and mDeBERTa). We suggest and employ recommendations to tackle these challenges.
Our results reveal low performance, even for English, and severe overfitting. We analyse various factors that influence these results, including label ambiguity, label set and dataset size. We find that the IQA performance is poor in high- (English, German) and low-resource languages (Bavarian) and that it is beneficial to have a large amount of training data. 
Further, GPT-4o-mini does not possess enough pragmatic understanding to generate high-quality IQA data in any of our tested languages. %
 \\ \newline  }

\begin{document}

\maketitleabstract

\section{Introduction}\label{sec:introduction}
Indirect Question Answering (IQA) is about recognizing the polarity of indirect answers, as illustrated in Table \ref{tab:sentence_samples}. 
Our motivation to study the task in both standard and non-standard languages lies in the practical relevance of IQA for communication with digital assistants, where users often interact in everyday, non-standard varieties. 
As \citet{blaschke2024dialectspeakers} show, German dialect speakers have interest in using conversational technologies in their native language -- for which indirectness is as relevant as it is common. For example,~\citet{sanagavarapu2022swdaia} find 77\,\% of 5-minute telephone conversations contain at least one yes/no question with an indirect answer. However, a large number of NLP (Natural Language Processing) tasks that are relevant for digital assistants are primarily researched in English.
In contrast, Bavarian is a low-resource, non-standardized German dialect.
It has recently been the subject in several NLP studies, including for tasks like slot and intent detection \cite{vandergoot2021xsid, winkler2024sid, krückl2025sid, blaschke-etal-2026-standard}, which can be pragmatically demanding.
However, no IQA datasets for Bavarian exist yet.

In general, IQA is understudied, especially in the multilingual or low-resource domain~(\S\ref{sec:related_work}).
Performance of limited previous work systems is usually rather poor and not up-to-par with the performance of tasks like POS tagging in monolingual \cite{awwalu2020pos, li2022pos, mohammad2025pos}, multilingual \cite{snyder2008multilingualpos, naseem2009multilingualpos, plank2016pos} and low-resource \cite{Mitri2025POS, haq2025pos, deore2025poslowresource} settings, or natural language inference \cite{bowman2015nli, storks2020nli, ning2025nli}.

Although the performance of IQA is not outstanding even in standard languages like English, the qualitative challenges underlying the task are rarely discussed beyond quantitative scores. Yet especially for tasks requiring deep pragmatic understanding, non-trivial language phenomena can introduce unexpected difficulties for modelling. %

\begin{table}[]
    \centering
    \resizebox{0.47\textwidth}{!}{%
    \begin{tabular}{cll}
        \toprule
         \flaguk & I'm early? & A little. \\
         \flagde & Komm ich zu früh? & Ein bisschen. \\
         \flagba & Bin i z'fria? & A bissal. \\ \bottomrule
    \end{tabular}%
    }
    \caption{Parallel example sentences from \inqaplus\ in all three available languages. The indirect answers correspond to \textit{Yes}.}
    \label{tab:sentence_samples}
\end{table}

\paragraph{Contributions}
\begin{itemize}
    \item We release two datasets, \inqaplus~(\S\ref{subsec:inqaplus}) and \geniqa~(\S\ref{subsec:geniqa}),\footnote{\repo} in English, German and Bavarian with pairs of questions and indirect answers featuring natural translations on the one hand and artificial data generated by GPT-4o-mini \cite{openai2024gpt} on the other.
    \item We experiment with and test the effect of data quality and quantity~(\S\ref{subsec:evaluation_iqa_experiments}) and demonstrate how to alleviate challenges arising during the experimentation.
    \item We analyse and discuss the implications of indirectness~(\S\ref{sec:discussion}) in an overview of the IQA task and why it is so difficult for humans and state-of-the-art large language models (LLMs) alike~(\S\S\ref{subsec:inqaplus_iaa} and \ref{subsec:geniqa_annotation_accuracy}).
\end{itemize}

\begin{table*}[]
    \centering
    \begin{tabular}{lll}
        \toprule
         \makecell[l]{\textbf{1 - Yes}\\This is Long Tieng, right?\\\textit{Right.}} & \makecell[l]{\textbf{2 - No}\\You hungry?\\\textit{Just black coffee for me.}} &  \makecell[l]{\textbf{3 - Conditional Yes}\\Will you be home tonight?\\\textit{I will be if you want me to be.}} \\ \midrule
         \makecell[l]{\textbf{4 - Neither yes nor no}\\Is anyone still in there?\\\textit{I don't know.}} & \makecell[l]{\textbf{5 - Other}\\Another trick question, right?\\\textit{Open the window.}} & \makecell[l]{\textbf{6 - Lacking context}\\Hey, you guys want sangria?\\\textit{It's Guys' Night!}} \\
         \bottomrule
    \end{tabular}
    \caption{Example questions and \textit{answers} from \inqaplus\ illustrating each \textbf{label}'s meaning.}
    \label{tab:label_examples}
\end{table*}

\section{Related Work}\label{sec:related_work}
Current research seldomly reflects on the specific challenges of IQA, especially the role of indirectness. Since understanding implicit messages relies on good pragmatic capabilities, dataset design can strongly influence model performance. The following related works not only include available resources but also highlight the diverse challenges of indirectness in NLP tasks.

\paragraph{IQA resources are good, but sparse.}
Even in English, IQA is one of the lesser researched tasks, with three notable data resources available: Circa \cite{louis2020circa}, Friends-QIA \cite{damgaard2021friendsqia} and SwDA-IA \cite{sanagavarapu2022swdaia}. Friends-QIA and SwDA-IA draw their data from existing resources like TV show scripts or phone call transcriptions, whereas Circa was created by large-scale crowdsourcing. The datasets consist of polar questions with one or multiple indirect answers in English which are annotated with labels denoting the polarity of the indirect answer. 

Multilingual IQA work is even more rare: \citet{wang2023indirectanswers} created a high-quality but unavailable multilingual corpus for 8 languages. IndirectQA EN \cite{mueller2024indirectqa} provides the first publicly available parallel multilingual dataset in English, French and Spanish. The data stems from parallel subtitles from the opensubtitles v2018 corpus\footnote{\url{http://www.opensubtitles.org/}} \cite{lison2016opensubtitles2018} and is also hand-annotated. It serves as the base of \inqaplus\ (\S\ref{subsec:inqaplus}). In our experiments, we replicate the experiments of \citet{mueller2024indirectqa}\footnote{\citet{mueller2024indirectqa} compare the IQA performance across test genres (comedy vs.\ crime). We do not see any distinctive performance differences between the genres and provide our per-genre results in Appendix \ref{app:inqa_genres}.} and get close results.

Analysing the IQA performance in previous work grants insights into possible challenges and performance to be expected. All datasets use 5 to 9 labels for classification. For IQA, we observe from prior research that, for example, the size of the context around the indirect answer matters for the performance \cite{louis2020circa, damgaard2021friendsqia}. More context can increase the performance of both machines and human annotation. \citet{paci2025llmimplicity} show this for implicit message understanding in an Italian politics domain, which could inform IQA, too. 

\paragraph{Indirectness is diverse.}
Indirectness is involved in a variety of NLP tasks, most prominently in sarcasm detection \cite{zhang2021sarcasm, jayaraman2022sarcasm, khan2025sarcasmlowresource}, implicit message understanding \cite{paci2025llmimplicity} and IQA (cf. previous paragraph). 
Because indirectness is a common occurrence in everyday human communication, it is prevalent in  the real-world.

Indirectness can challenge model performance, as \citet{paci2025llmimplicity} underscore with their work on implicit political message understanding: indirectness occurs frequently and many models struggle with the required pragmatic understanding.

\section{\inqaplus: Hand-Curated Test Dataset}\label{subsec:inqaplus}
\inqaplus\ is our multilingual extension of IndirectQA (\citealp{mueller2024indirectqa}; \S\ref{sec:related_work}). It serves as a small, yet high-quality test dataset. 
It consists of 438 hand-annotated question-answer pairs sourced from the opensubtitles v2018 \cite{lison2016opensubtitles2018} corpus (examples in Table~\ref{tab:label_examples}). The size is derived from the French and Spanish data splits of IndirectQA. The question answer pairs are parallel in English, Standard German and Bavarian. 
\inqaplus\ is completely new and separate from IndirectQA. The English sentences are distinct from the sentences in IndirectQA because only a few sentences from IndirectQA were parallel with the German opensubtitles v2018 \cite{lison2016opensubtitles2018} corpus, so no direct extension to IndirectQA could be produced. 
The data statement can be found in Appendix \ref{app:inqaplus_statement}.

As motivated in \S\ref{sec:introduction}, we want to contribute more Bavarian data resources to the NLP space. English as the highest resource language builds the baseline for our experiments, whereas Standard German rounds off our selection as another well-researched standard language that is closely related to Bavarian.

To find suitable candidate pairs, we lightly pre-filter the raw opensubtitles v2018 corpus, that is parallel between English and German, for questions that are followed by a line that does not contain the direct answer particles \textit{Yes} or \textit{No}. We only select indirect answers that sound plausible as an answer to the preceding question and that do not contain any direct answer particles including colloquial ones like \textit{Yeah} or \textit{Nope} (also see Appendix~\ref{app:inqaplus_statement}). 
We manually label the data with the label set described in \S\ref{subsec:annotations_and_label_defs} and show the label distribution in Table~\ref{tab:label_distributions}.

\subsection{Annotations and label definitions}\label{subsec:annotations_and_label_defs}
We use the same label set as IndirectQA \cite{mueller2024indirectqa} to maintain data compatibility.
The labels are defined as follows:

\begin{enumerate}
    \item \labelone: A clear yes or all gradients of yes (including weaker forms, e.g., maybe yes). %
    \item \labeltwo: A clear no or all gradients of no.
    \item \labelthree: A yes that only holds if certain conditions are true.
    \item \labelfour: A neutral answer that lies in the middle of yes and no.
    \item \labelfive: The sentence does not match the questions as an answer.
    \item \labelsix: Without further context, the answer cannot be clearly categorized.
\end{enumerate}

\noindent
We illustrate the meaning of each label further with the examples in Table \ref{tab:label_examples}. 
The label set is a modification from the six labels of \citet{louis2020circa} and \citet{damgaard2021friendsqia}. \labelsix\ replaces the \textit{N/A} label from \citet{damgaard2021friendsqia}. Although it sounds similar, it is different from \labelfour\ and \labelfive: Answers labelled as \labelsix\ have a clear polarity, unlike \labelfour, and does function as an answer the question, unlike \labelfive, only the position on the scale from yes to no is unclear without further context.

For pragmatically difficult cases, like questions with negations (negative questions), we set additional guidelines to ensure uniform annotations. 
We follow polarity-based annotation \cite{holmberg2012syntax, holmberg2013syntax}, focusing on the grammatical form of the question (e.g., positive or negative wording) rather than the polarity of its meaning.
For example, the question \textit{Are you not Moby?} from \inqaplus\ can be answered with either \textit{(Yes) I am Moby} or \textit{(No) I am not Moby}, leading to \labelone\ or \labeltwo\ annotations respectively. In contrast, the answer \textit{Yes, I am not Moby} falls under truth-based polarity \cite{kuno1975structure, holmberg2012syntax, holmberg2013syntax}.
Appendix \ref{app:inqa_annotation_details} contains more detailed annotation guidelines and examples.
 
\paragraph{Label set variations} 
To investigate the effect of the label set complexity, we conduct experiments with two simplified versions of the label set. 

In the \textbf{\remapped} version, we merge labels that are similar to each other: firstly, \labelone\ and \labelthree\ (which both denote instances that signify gradients of \labelone), and secondly, \labelfive\ and \labelsix\ (which contain samples that are unclear in their function as an answer). This creates sharper distinctions between clear and vague answers.
\remapped\ consists of four labels.

The \textbf{\yesno}\ version only contains instances with the labels \labelone\ and \labeltwo.
This is the most minimal version of IQA. 
For \inqaplus, we omit all instances with other labels (leaving 283 instances). 
For \geniqa, we generate a dedicated \yesno\ version in each language to preserve the dataset size of 1,500.

We report results for these label set variations in~\S\ref{subsec:simplifying_label_set} and discuss their benefits and drawbacks in~\S\ref{sec:discussion}.

\begin{table}[]
    \centering
    \resizebox{\linewidth}{!}{%
    \begin{tabular}{c|crrr}
        \toprule
          & \multicolumn{1}{c}{\inqaplus} &  \multicolumn{3}{c}{\geniqa} \\
        Label & \flagmulti & \flaguk & \flagde & \flagba \\ \midrule
        \labelone & 183 & 400 & 250 & 218 \\
        \labeltwo & 100 & 132 & 210 & 182 \\
        \labelthree & 18 & 319 & 274 & 306 \\
        \labelfour & 79 & 477 & 464 & 468 \\
        \labelfive & 34 & 71 & 141 & 191 \\
        \labelsix & 24 & 101 & 161 & 135 \\ \midrule
        Total & 438 & \multicolumn{3}{c}{1.500}\\
        \bottomrule
    \end{tabular}
    }
    \caption{Label distributions of \inqaplus\ and \geniqa. The distribution of \inqaplus\ applies to all languages due to the parallelism.}
    \label{tab:label_distributions}
\end{table}

\subsection{Inter-annotator agreement}\label{subsec:inqaplus_iaa}
IQA is a pragmatically hard task. We derive this not only from the statements of previous work
\cite{damgaard2021friendsqia, mueller2024indirectqa}, but also from our finding that humans and machines alike find it hard to classify indirect answers as the Inter-Annotator Agreements (IAA) shows.

\begin{figure}
    \centering
    \includegraphics[width=\linewidth]{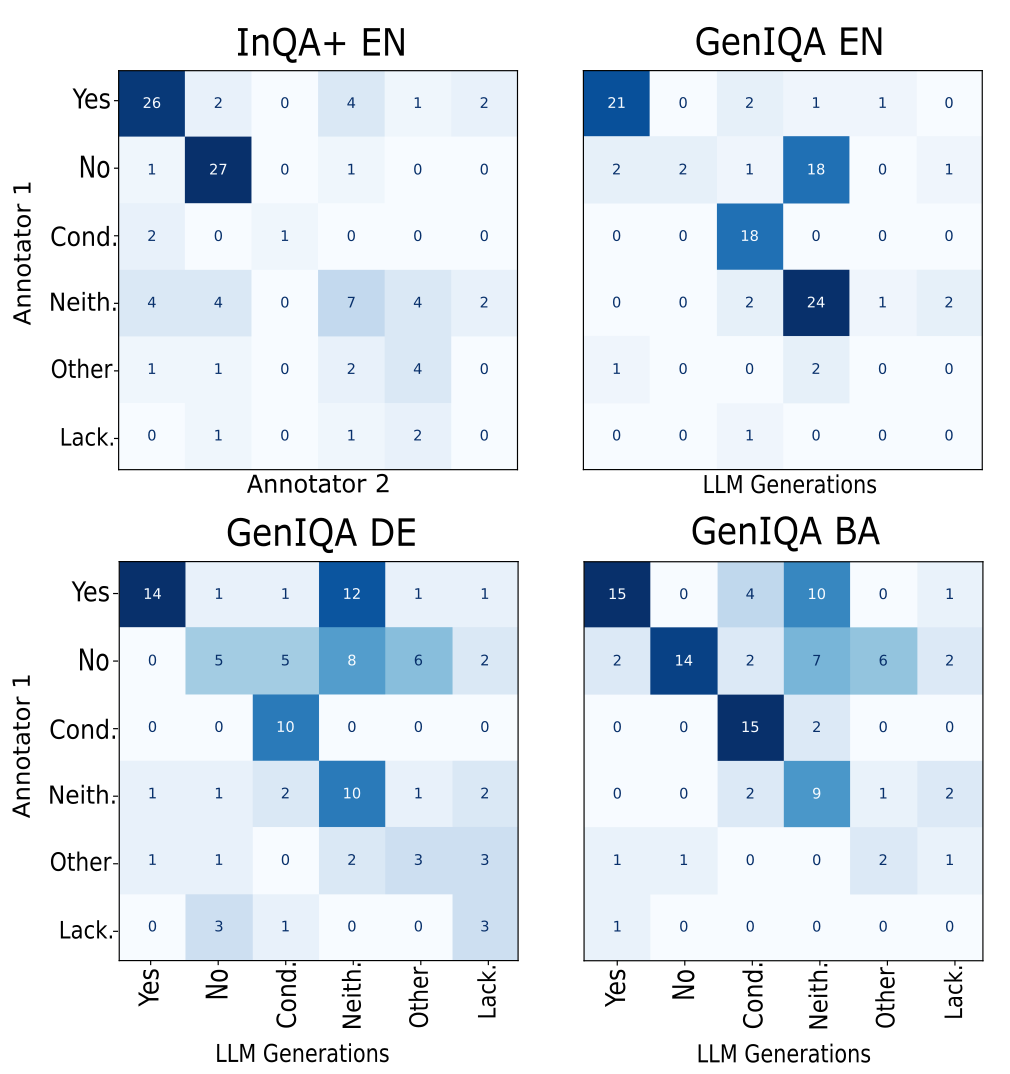}
    \caption{Confusion matrices between two annotators (top left) and the \geniqa\ labels as originally generated vs.\ re-annotated by the main annotator, respectively on 100 sentences from each dataset. Cond. = Conditional Yes; Neith. = Neither Yes nor No; Lack. = Lacking Context.}
    \label{fig:iaa_matrices}
\end{figure}

In our hand-labelled \inqaplus\ data, we obtain a moderate agreement with a Cohen's Kappa of 0.53 on a sample of 100 \inqaplus\ EN instances which were double-annotated by expert annotators.  In other IQA resources, we observe similarly moderate agreements, for example in IndirectQA \cite{mueller2024indirectqa} with a Cohen's Kappa of 0.54 on 205 double-annotated pairs and Circa \cite{louis2020circa} with a Fleiss' Kappa of 0.61. %

Figure~\ref{fig:iaa_matrices} (top left) shows that we have the most agreement for cases that are either clearly \labelone\ or \labeltwo\ and low agreement for every other label.
Our IAA is $\kappa=0.57$ for \remapped\ and $\kappa=0.70$ for \yesno.

\section{\geniqa: Artificial Training Dataset}\label{subsec:geniqa}
As the availability of IQA data is limited, especially for low-resource languages, we experiment with LLM-generated training data.
We create \geniqa, which consists of 1,500 question-answer pairs which we generate with GPT-4o-mini \cite{openai2024gpt} in English, Standard German and Bavarian. The data statement can be found in Appendix \ref{app:geniqa_statement}. All languages were generated independently and not translated. The pairs were annotated by the model at generation time with the same label set as \inqaplus\ (\S\ref{subsec:annotations_and_label_defs}). The label distribution per language is found in Table \ref{tab:label_distributions}.

As preliminary experiments showed high language quality for generated English and Standard German texts for many LLMs, we focus on Bavarian for selecting an LLM to generate \geniqa, as it is a priori less clear which LLM works for Bavarian. We perform experiments on dialect generation with 9 LLMs\footnote{gemma-2-2b-it \cite{gemmateam2024gemma2}, Llama-3.2-1B-Instruct \cite{meta2024llama}, OLMo-2-1124-7B-Instruct \cite{olmo20242olmo2furious}, GPT-4o-mini \cite{openai2024gpt}, LLäMmlein\_1B \cite{pfister2024llammlein}, Betzerl \cite{caidas2025betzerl} and two sizes of EuroLLM \cite{martins2025eurollm} (EuroLLM-1.7B-Instruct and EuroLLM-9B-Instruct), more details about the LLM dialect experiments and prompt tuning is provided in Appendix \ref{app:llm_generation_tests}.} to find a model that can generate Bavarian dialect.
GPT-4o-mini was capable of generating the best dialect output of all tested models, albeit still not of high quality as we discuss in the next Section. 
For comparability between languages, we use the same prompt to also use GPT-4o-mini to generate the English and Standard German datasets. %
The prompt includes six examples, one per label (more details in Appendix \ref{app:llm_generation_tests}). We generate 50 question-answer pairs per prompt until we reach 1,500. We remove duplicates and need 46 iterations for \geniqa\ EN, 34 for DE and 33 for BAR.

\paragraph{Dataset variations}
As with \inqaplus, we also experiment with reduced label set variations (\remapped, \yesno; \S\ref{subsec:annotations_and_label_defs}). To compare the performance of IndirectQA EN \cite{mueller2024indirectqa} and \geniqa\ trained models on a more equal basis, we perform experiments with a sized down version of \geniqa\ EN: \textbf{\geniqa\ EN \textsc{small}}. Like IndirectQA EN, it consists of 615 instances and exhibits a similar label distribution to create comparable conditions.

\subsection{Bavarian \geniqa\ language quality}\label{subsec:geniqa_dialect_quality}
Generally, the quality of the generated dialect output is low and the LLMs we tested are not capable of generating authentic and natural-sounding dialect. 
We investigate the quality of the generated data both manually and by surveying a larger set of dialect speakers.

Manual inspections reveal that many instances of the Bavarian \geniqa\ dataset either only contain one dialect word, use auxiliary verbs incorrectly or choose the wrong genus of the determiner. Furthermore, the dialect is not generated consistently. In a sample of 100 instances, 70\,\% of questions and 6\,\% of answers were Standard German, as classified by a native dialect speaker. 
We speculate that a possible reason could be that questions are much rarer than indicative statements in the pre-training data for GPT-4o-mini, and questions in a low-resource dialect might be especially rare.
The rest of the data contained at least some nuances of dialect. 
The generated dialect also often resembles colloquial German more than genuine Bavarian.
Nevertheless, some instances contain generations of correct characteristic Bavarian grammar phenomena like clitics, the merging of the word \textit{es} (ENG. it) into the preceding words, like \textit{is's} (DEU. ist es; ENG. is it). We provide more insights into the manually assessed quality in Appendix \ref{app:generated_dialect_quality}.

We confirm the poor quality we observed by surveying native speakers. 
The survey was hosted via SoSci Survey \cite{leiner2025sosci} and distributed via various social channels including Discord, Whatsapp and Reddit. 76 active dialect speakers, the majority stemming from the Upper and Lower Bavarian regions, answered the survey and assessed ten indirect answers on whether they include dialect and if they find the answer authentic in that they themselves would produce such a sentence; nine of those answers are LLM-generated, one is a hand-translated instance from \inqaplus. 

When mapping the scales to numbers (0\,=\,worst, 3\,=\,best), the LLM-generated answers of GPT-4o-mini received a dialectness rating of 1.1 and a authenticity rating of 1.4 on average (compared to the hand-translated instance with ratings of 2.5 and 1.8),\footnote{%
The ratings are higher when only considering answers by respondents from the region that the translator is from (refer to Figure \ref{fig:survey_dialect_auth_lower_higher} in Appendix \ref{app:generated_dialect_quality} for more details).
}
confirming the low quality. However, we do not strife to create a high-quality artificial dataset. Instead our \geniqa\ experiments demonstrate the current state of LLMs with regards to the Bavarian dialect in a naive and intuitive approach. The full questionnaire, participant information and results can be found in Appendix~\ref{app:generated_dialect_quality}.

\subsection{Labelling accuracy}\label{subsec:geniqa_annotation_accuracy}
We manually re-annotate 100 instances of \geniqa. Our \geniqa\ datasets only reach annotation accuracies of 65\,\% for EN, 45\,\% for DE and 55\,\% for BAR. The overall quality is low with the models performing best on English generations. LLMs are good at generating coherent texts that fulfil social expectations. However, as soon as more complex phenomena like indirectness or references are involved, they struggle \cite{hu2023finegrainedcomparisonpragmaticlanguage, ma2025pragmaticseralargelanguage}. \citet{qiu-etal-2023-chatgpt} found that ChatGPT does not process implicatures like humans do, because it cannot switch between pragmatic and semantic interpretations easily.

The label distributions differ across the language splits, as visualised in Figure~\ref{fig:iaa_matrices}. For each split, the labels \labelone, \labelthree\ and \labelfour\ have the highest accuracy with respect to the manual annotations, unlike the IAA between two humans, who have high agreement for \labeltwo\ as well.

\section{Experimental Setups}\label{sec:experimental_setup}
We explore the effect of data quantity and quality with three fine-tuneable multilingual models in their base sizes for comparability: mBERT \cite{devlin2019bert} is a widely used baseline for many experiments in a lot of research papers, for example in the related Circa \cite{louis2020circa} and IndirectQA \cite{mueller2024indirectqa} research, sarcasm research of \citet{jayaraman2022sarcasm} and \citet{zhang2021sarcasm} and Bavarian tasks like slot and intent detection \cite{vandergoot2021xsid,winkler2024sid}. We complement it with XLM-RoBERTa (XLM-R) \cite{conneau2020xmlroberta} and mDeBERTa \cite{he2020mdeberta}, which are used in other Bavarian research by \citet{peng2024barner} and \citet{krueckl2025xsid}. mBERT is the only model which has seen Bavarian training data (from Wikipedia) during pre-training.\footnote{\url{https://github.com/google-research/bert/blob/master/multilingual.md}}
We refrain from using instruction-tuned GPT models as we saw low label generation quality (\S\ref{subsec:geniqa_annotation_accuracy}), and because fine-tuned encoder models often lead to better classification results than prompting instruction-tuned models \cite{mosbach-etal-2023-shot, qorib-etal-2024-decoder, weller2025seqvsseqopen}.

We have multiple training setups to test the effects of data quality, language and quantity in different scenarios. In the first setup~(\S\ref{subsec:standard_results}), we compare training on manually vs.\ artificially created data. 
In further experiments, we vary the size of the training split~(\S\ref{subsec:results-size}) by adding Friends-QIA \cite{damgaard2021friendsqia} to the training data and use the dataset versions with the reduced label sets~(\S\ref{subsec:simplifying_label_set}). The datasets' annotations are compatible without further processing. The data is stored in text files with one question, indirect answer and label per line, separated by a tab space.

\begin{table}[]
    \centering
    \resizebox{0.9\linewidth}{!}{
    \begin{tabular}{@{}l|cc@{}}
        \toprule
        Parameter & Grid Seach & Random Search \\ \midrule
        Learning Rate & [1e-4, 1e-5, 1e-6] & 1e-6–-5e-4 \\
        Batch Size & [4, 8, 16, 32] & [4, 8, 16, 32] \\
        Warm-up Ratio & [0.1, 0.3, 0.5] & 0.1–-0.5 \\
        Weight Decay & [0.001, 0.01, 0.1] & 0.001–-0.1 \\
        Dropout Rate & [0.1, 0.3, 0.5] & 0.1–-0.5 \\
        Label Smoothing & [0.1, 0.3, 0.5] & 0.05–-0.5 \\ 
        \bottomrule
    \end{tabular}
    }
    \caption{Hyperparameter search spaces for grid and random search.}
    \label{tab:parameter_search_spaces}
\end{table}

For each training setup, we fine-tune the hyperparameters learning rate, batch size, warm-up ratio, weight decay, dropout rate and label smoothing in the scope of the search spaces in Table \ref{tab:parameter_search_spaces} individually. 
We choose hyperparameter values based on accuracy.
The optimal hyperparameters differ across datasets, label set variations and language models. We explain the fine-tuning method and details in Appendix \ref{app:hyperparameter_finetuning}. We employ grid and random search to find the best set of parameters. We observe that the grid-searched parameters yield better dev accuracy than the random-searched ones. Additionally, to avoid overfitting as much as possible, we employ early stopping in training.

We train each model on three random seeds and report mean accuracy, macro-F1 and standard deviations, as the scores vary greatly per seed and single runs are unreliable and noisy. We transparently show the capabilities of SOTA models and discuss them with regards to high deviation.

\begin{table*}
    \centering
    \resizebox{\textwidth}{!}{
    \begin{tabular}{@{}l|ccc}
    \toprule
        & \multicolumn{3}{c}{Accuracy} \\ \midrule
         & \makecell[c]{\inqaplus\ EN} & \makecell[c]{\inqaplus\ DE} & \makecell[c]{\inqaplus\ BAR}  \\ \midrule
        Majority Class Baseline & 41.78 & 41.78 & 41.78 \\ [1ex]
        mBERT trained with & & &  \\
        \qquad IndirectQA EN  & 34.02 \stdev{2.57} & 36.30 \stdev{0.60} & 38.58 \stdev{3.59} \\
         \qquad \geniqa\ EN  & 27.25 \stdev{2.87} & 23.29 \stdev{7.15} &  21.31 \stdev{3.49} \\
         \qquad  \geniqa\ DE  & 20.09 \stdev{4.56} & 19.18 \stdev{3.14} &  20.55 \stdev{1.95} \\
         \qquad  \geniqa\ BAR & 16.59 \stdev{0.92} & 15.37 \stdev{1.99} & 18.87 \stdev{1.99} \\ [1ex]
         mDeBERTa & & &  \\
        \qquad IndirectQA EN  & 33.26 \stdev{7.42} & 32.65 \stdev{8.58} &  36.45 \stdev{2.12} \\
         \qquad \geniqa\ EN  & 33.11 \stdev{9.76} & 32.04 \stdev{10.9\rlap{7}} &  30.21 \stdev{11.5\rlap{6}} \\
         \qquad  \geniqa\ DE  & 25.57 \stdev{7.49} & 25.42 \stdev{7.05} &  22.98 \stdev{2.42} \\
         \qquad  \geniqa\ BAR  & 20.02 \stdev{3.47} & 17.88 \stdev{3.58} & 20.47 \stdev{1.94} \\ [1ex]
        XLM-R & & & \\
        \qquad IndirectQA EN  & 39.42 \stdev{2.30} & 41.25 \stdev{0.57} &  40.18 \stdev{1.81} \\
         \qquad \geniqa\ EN  & 32.88 \stdev{3.36} & 29.60 \stdev{3.90} &  27.63 \stdev{5.48} \\
         \qquad  \geniqa\ DE  & 19.71 \stdev{4.03} & 21.00 \stdev{3.59} &  22.60  \stdev{3.90}\\
         \qquad  \geniqa\ BAR  & 22.15 \stdev{3.84} & 22.37 \stdev{1.95} & 22.83 \stdev{2.20} \\ 
    \bottomrule
    \end{tabular}%
    \begin{tabular}{|l|ccc@{}}
    \toprule
        & \multicolumn{3}{c}{Macro F1} \\ \midrule
         & \makecell[c]{\inqaplus\ EN} & \makecell[c]{\inqaplus\ DE} & \makecell[c]{\inqaplus\ BAR}  \\ \midrule
         & 9.82 & 9.82 & 9.82 \\ [1ex] %
         & & &  \\ %
          & 15.23 \stdev{1.87} & 14.51 \stdev{3.05} & 13.42 \stdev{1.68}  \\
           & 16.69 \stdev{2.39}  &  \phantom{0}9.85 \stdev{6.38} & \phantom{0}7.60 \stdev{2.63}  \\
           & 12.41 \stdev{6.19}  & 13.33 \stdev{4.36}  & 10.35 \stdev{3.29}  \\
           & 12.01 \stdev{0.84}  &  15.67 \stdev{2.48} & 16.16 \stdev{2.08} \\ [1ex]
           & & &  \\ %
          & 12.93 \stdev{2.67} & 12.83 \stdev{2.84} &  11.71 \stdev{0.98} \\
           &  19.27 \stdev{3.93} &  18.56 \stdev{4.44} &  16.61 \stdev{4.44} \\
           &  16.90 \stdev{2.70} &  17.40 \stdev{3.99} &  13.11 \stdev{4.13} \\
           &  20.15 \stdev{4.65} &  18.01 \stdev{3.79} & 18.22 \stdev{2.57} \\ [1ex]
         & & & \\ %
          &  14.90 \stdev{4.63} & 15.05 \stdev{4.87} &  12.16 \stdev{2.10} \\
           &  20.43 \stdev{2.12} &  18.04 \stdev{2.54} &  16.80 \stdev{2.83} \\
           &  16.56 \stdev{3.34} &  18.03 \stdev{2.46} &  15.56 \stdev{4.17} \\
           &  21.35 \stdev{4.44} &  21.55 \stdev{2.49} & 19.29 \stdev{0.68} \\ 
    \bottomrule
    \end{tabular}%
    }
    
    \caption{Average accuracy and macro F1 scores over three seeds and standard deviations of models trained on our datasets, evaluated on \inqaplus. The train/dev split follows a ratio of 80-20. }
    \label{tab:results_std}
\end{table*}

\section{Results and Analysis}\label{subsec:evaluation_iqa_experiments}
Since the IQA task is challenging, we focus our experiments on the effects of data quality and quantity to see which is more influential to reduce overfitting. For the analysis of a difficult task, it is important to take multiple metrics into account, in our case: accuracy and F1 scores. The performance cannot always be read directly from the accuracy, and the combination with F1 scores reveals more model capabilities.

\begin{table*}
    \centering
    \resizebox{\textwidth}{!}{
    \begin{tabular}{@{}l|ccc@{}}
    \toprule
        & \multicolumn{3}{c}{Accuracy} \\ \midrule
         & \makecell[c]{\inqaplus\ EN} & \makecell[c]{\inqaplus\ DE} & \makecell[c]{\inqaplus\ BAR}  \\ \midrule
        Majority Class Baseline  & 41.78 & 41.78 & 41.78 \\ 
        Majority Class Baseline \remapped & 45.89 & 45.89 & 45.89 \\ 
        Majority Class Baseline \yesno & 64.66 & 64.66 & 64.66 \\ [1ex]
        mBERT trained with & & &  \\
        \qquad IndirectQA EN \remapped & 27.02 \stdev{4.53} & 23.36 \stdev{8.95} & 23.44 \stdev{11.46} \\
         \qquad IndirectQA EN + Friends-QIA \remapped & 47.56 \stdev{2.91} & 40.56 \stdev{8.59} &  34.25 \stdev{7.46} \\ [1ex]
        \qquad IndirectQA EN \yesno & 63.96 \stdev{3.59} & 60.42 \stdev{8.28} &  60.42 \stdev{8.28} \\
         \qquad \geniqa\ EN \yesno & 62.78 \stdev{3.34} & 68.43 \stdev{0.82} &  64.78 \stdev{0.74} \\
         \qquad  \geniqa\ DE \yesno & 51.59 \stdev{10.41} & 63.02 \stdev{4.42} &  59.60 \stdev{4.22} \\
         \qquad  \geniqa\ BAR \yesno & 65.37 \stdev{0.93} & 65.96 \stdev{1.08} & 64.19 \stdev{0.41} \\ [1ex]
         \qquad IndirectQA EN (repeated from Table~\ref{tab:results_std}) & 34.02 \stdev{2.57} & 36.30 \stdev{0.60} & 38.58 \stdev{3.59} \\
         \qquad Friends-QIA & \quad 43.07 \stdev{4.80} & 35.77 \stdev{8.17} &  30.29 \stdev{6.90} \\
         \qquad IndirectQA EN + Friends-QIA & \quad 46.27 \stdev{1.94} & 38.81 \stdev{2.25} &  31.74 \stdev{5.27} \\
         \qquad \geniqa\ EN \textsc{small} & 30.82 \stdev{5.31} & 27.25 \stdev{8.33} & 24.51 \stdev{4.82} \\ 
    \bottomrule
    \end{tabular}%
    \begin{tabular}{|l|ccc}
    \toprule
        & \multicolumn{3}{c}{Macro F1} \\ \midrule
         & \makecell[c]{\inqaplus\ EN} & \makecell[c]{\inqaplus\ DE} & \makecell[c]{\inqaplus\ BAR}  \\ \midrule
         & 9.82 & 9.82 & 9.82 \\
         & 15.73 & 15.73  & 15.73  \\
         & 39.27 & 39.27  & 39.27  \\ [1ex]
         & & &  \\ %
         & 17.10 \stdev{2.54} & 14.01 \stdev{6.49} & 13.48 \stdev{7.04} \\ %
         & 35.40 \stdev{2.46} & 30.13 \stdev{5.69} & 24.21 \stdev{3.96} \\ [1ex] %
          & 51.26 \stdev{10.45} & 44.47 \stdev{5.88} & 44.38 \stdev{5.74}  \\ %
         & 50.25 \stdev{7.85} & 58.01 \stdev{6.83} &  46.42 \stdev{7.32} \\ %
        & 48.81 \stdev{12.30} & 61.68 \stdev{3.48} &  55.25 \stdev{2.60} \\ %
        & 43.80 \stdev{6.93} & 56.87 \stdev{3.64} & 58.30 \stdev{0.54} \\ [1ex] %
        & 15.23 \stdev{1.87} & 14.51 \stdev{3.05} & 13.42 \stdev{1.68}  \\
        & 21.92 \stdev{1.41} & 17.75 \stdev{3.81} & 14.16 \stdev{1.29} \\ %
         & 23.43 \stdev{0.89} & 19.76 \stdev{0.63} & 16.27 \stdev{2.14} \\  %
        & 16.64 \stdev{1.92}  & 11.48 \stdev{2.80}  & 10.60 \stdev{2.54}  \\ %
    \bottomrule
    \end{tabular}%
    }
    \caption{Average accuracy and macro F1 scores over three seeds and standard deviations of models trained on experimental dataset variations. The \remapped\ and \yesno\ setups are evaluated on the corresponding \remapped\ and \yesno\ \inqaplus\ test sets.}
    \label{tab:results_variants}
\end{table*}

\subsection{Data quality versus quantity }\label{subsec:standard_results}
We first present our main investigation: %
comparing small high-quality training data (IndirectQA EN; \citealp{mueller2024indirectqa}) against larger, but low-quality artificial training data (\geniqa) to research the trade-off of high quality versus high quantity.

\paragraph{Using manual versus generated data}
When analysing the results of mBERT in Table~\ref{tab:results_std}, we observe a major issue: all scores lie below the majority baseline accuracy (41.8\,\%) due to overfitting and the F1 scores are low, even if they almost always surpass the majority class baseline F1.

The highest accuracy score (38.6\,\%) is achieved by the model trained with IndirectQA EN when evaluated on the Bavarian \inqaplus\ test set. This score is close to the majority class baseline due to a bias for the majority class, revealing that the training was not effective, which the low F1 score (13.4\,\%) also indicates. 
In previous work, we see similar behaviour and discuss this in \S\ref{sec:discussion}.

From the mBERT results, the models trained with the artificial but larger \geniqa\ datasets have lower accuracy than the IndirectQA EN models, but the higher diversity in the data (stemming from the larger dataset size) benefits the generalisation performance on the in-language test data.

The best F1 scores for the test sets stem from the models trained on their respective \geniqa\ training dataset (16.7\,\% for \inqaplus\ EN and 16.2\,\% for \inqaplus\ BAR). The only exception to this is \inqaplus\ DE, where the model trained with \geniqa\ BAR reaches the best performance.
This is likely due to the generated ``Bavarian'' data largely resembling German~(\S\ref{subsec:geniqa_dialect_quality}).

With mDeBERTa \cite{he2020mdeberta} and XLM-R \cite{conneau2020xmlroberta},
results only approach the majority class baseline (highest score: 41.3\,\% accuracy with XLM-R). 
While accuracy of \geniqa\ trained models lags behind the performance of IndirectQA EN, mDeBERTa trained on \geniqa\ EN performs just as well (accuracy) or even better (F1) than IndirectQA EN. XLM-R produces the highest accuracies for all languages when fine-tuned with the small IndirectQA EN. Further, we see large increases in the F1 scores when fine-tuning with any \geniqa\ dataset, 
indicating that the larger dataset size might be beneficial for learning class representations beyond the majority class.
Many of highest F1 scores are produced by models trained on the Bavarian version of \geniqa. This might be due to that split having a more uniform label distribution than the English and German splits (Figure~\ref{fig:iaa_matrices}).

As the overall performance is already unsatisfactory in English with no accuracy scores surpassing the majority class baseline accuracy, we see even worse performance for Standard German and the low-resource dialect Bavarian (Table~\ref{tab:results_std}). In the standard setups, the small yet high-quality training data IndirectQA EN yielded the best accuracy scores for Standard German (32.7--41.3\,\%) and Bavarian (36.5--40.2\,\%). However, the F1 scores 
are again better with the larger \geniqa\ datasets, especially \geniqa\ BAR (on \inqaplus\ DE: 15.7--21.6\,\%; on \inqaplus\ BAR: 16.2--19.3\,\%). 

Because observed trends are similar across models, and to save compute, subsequent experiments are only with mBERT.

\paragraph{Performance on training and development sets}\label{subsec:train_and_dev_results}
Our greatest challenge is overfitting which occurred in each experimental setup. It becomes visible when comparing the accuracy scores of each model on the training and development sets.

All train (accuracy between 48.1--72.3\,\%) and dev scores (accuracy between 40.1--62.4\,\%) lie above the majority class baseline accuracy of 30.9--35.6\,\%, depending on the dataset. This shows that the training itself works. We provide the full results table and further discussion in Appendix \ref{app:train_dev_results}. That the scores are low even for English experiment setups underscores the task difficulty of IQA. 

The gaps of 2.2--12.4 percentage points (pps.) between train and dev accuracies reveal the grade of overfitting: the larger the gap, the more overfitting took place. We observe the largest discrepancies for XLM-R models fine-tuned with \geniqa\ EN and \geniqa\ BAR with gaps of 12.4 and 11.0 pps. between the train and dev set accuracy.

\subsection{Assessing the effect of dataset size}
\label{subsec:results-size}

Since we are interested in the effect of data quality versus quantity, we next experiment with datasets of varying sizes to specifically target characteristics like the data quantity. 
Enlarging the dataset size by adding data from different datasets can reveal what the original training data is lacking if the performance is not as expected.

We see this first when training with Friends-QIA \cite{damgaard2021friendsqia}, which is much larger than IndirectQA (5,930 versus 615) and which yields large accuracy gains over IndirectQA EN \cite{mueller2024indirectqa} (Table~\ref{tab:results_variants}) with an increase of 9.1 pps. on \inqaplus\ EN. The transfer capabilities to Standard German and Bavarian display lower accuracy (losses of 0.5--8.3\,pps.), but higher F1 scores
as the increases of 0.7--6.7\,pps. across all languages show. Training on a combined setup of Friends-QIA and IndirectQA EN pushes performance even further with increases of 1.5--3.2\,pps. over Friends-QIA alone. Thus, more data is beneficial for IQA in each language. However, at the same dataset size, the quality of the data is still more crucial for the performance, as the accuracy loss of \geniqa\ EN \textsc{small} (\S\ref{subsec:geniqa}) of 3.2-14.1\,pps. against IndirectQA EN reveals.

\subsection{Reduced ambiguity in label set}\label{subsec:simplifying_label_set}
Fusing and omitting labels can eliminate ambiguous labels like \labelsix\, which in many cases, as we observed, could have been resolved by hearing the intonations or seeing the mimics of the speaking actor in the movie, because humans usually distinguish literal meaning from sarcasm through intonations \cite{glenwright2014sarcasmnintonation}. Without context, it remains ambiguous.

By modifying the label sets, we expect better performance through task simplification and unambiguous labels, as the simplification already increased the human agreement (cf.~\S\ref{subsec:inqaplus_iaa}). For this reason, we experiment with models fine-tuned on the \remapped\ and \yesno\ dataset variants presented in \S\ref{subsec:annotations_and_label_defs}. We evaluate them on the corresponding \remapped\ and \yesno\ \inqaplus\ test sets. 

Altering the data can reduce label ambiguity. However, fusing the labels can also decrease the dataset size and obscure the granularity of the label set and lose nuances, which makes the annotations less informative. These operations can thus also have negative impact on the performance and need to be applied thoughtfully.

While we do not see improvements by training with \remapped\ alone (loss of 7~pps. accuracy in comparison to the full label variant, cf. Table \ref{tab:results_variants}), combining it with a \remapped\ version of Friends-QIA \cite{damgaard2021friendsqia} improves the performance clearly with F1 scores twice as high. We inspect this aspect of dataset size further in \S\ref{subsec:results-size}

The accuracy of \yesno\ lies around the majority class baseline. The F1 scores, however, are considerably higher than for the full label and \remapped\ training dataset variants. We observe increases of 34.1--51.9\,pps.\ over the majority class baseline F1, showing the simplicity of the minimalistic setup.

\section{Discussion and Learnings}\label{sec:discussion}

Both our experiments and previous work achieve results around the majority class baseline, confirming that IQA is highly challenging. Even in high-resource languages, the performance is low, and the gap widens in low-resource languages such as Bavarian.  
We thus share our learnings for IQA. 

\paragraph{Data quantity is important.}
A fundamental challenge we faced was a dataset that was too small for the pragmatically hard IQA task. We can confirm this with successful experiments using the larger combination of IndirectQA EN \cite{mueller2024indirectqa} and Friends-QIA \cite{damgaard2021friendsqia} (615 + 5,930 instances). Even \geniqa\ with a size of 1,500 instances is still too small to reach meaningful scores. 

From previous research, we deduce that a dataset of at least a size between $\sim$\,6,000 (Friends-QIA; \citealp{damgaard2021friendsqia}) and $\sim$\,35,000 (Circa; \citealp{louis2020circa}) might be necessary to learn enough pragmatic features from the training data.

Increasing the amount of data is not always possible, thus experimenting with various models is helpful as they each preprocess data differently. In our case, changing the model architecture alone does not improve the scores (\S\ref{subsec:standard_results}), at least not without also changing the model size. We see this from the research of \citet{sravanthi2024pub} who achieve results that even surpass human performance on response classification of answers with implied meaning with very large models like flan-t5-xxl \cite{chung2022flant5xxl} and llama-2-70b-chat \cite{meta2024llama}. As we established in \S\ref{sec:experimental_setup}, smaller sized encoder-LMs do not work well with IQA, which reflects in their performances during the experiments. 

\paragraph{Language interpretations and annotations are inconsistent.}
Indirectness is subjective and ambiguous.
As we discussed in \S\ref{subsec:inqaplus_iaa}, the annotations for IQA depend on individual interpretations of the indirect answers and the agreement is moderate (0.53). It suffers even more when the label set is ambiguous, meaning that the labels do not have clearly distinguishable boundaries and detailed annotation guidelines. Like this, the same labels can be applied differently in different datasets, which may lead to discrepancies in the annotations between the training data and test data, causing unexpected results. \S\S\ref{subsec:annotations_and_label_defs} and \ref{subsec:inqaplus_iaa} explain that such cases raise the need for clear annotation guidelines that can also help with the annotation reconciliation between data sources.

Many annotations will inevitably remain ambiguous and dependent on different perspectives. We want to note here that differing annotations are not errors per se, but can fall into the scope of human label variation \cite{plank-2022-problem}. Clear annotation guidelines help in avoiding ambiguities and annotation disagreements, which is why cleaning the data and correctly labelling it in the case of discrepancies may be important before the training and evaluation.

This is not a sole occurrence in IQA, but also persists in other pragmatically challenging tasks about indirect language, like sarcasm detection. The MUStARD (Multimodal Sarcasm Dataset; \citealp{castro2019mustard}) and CSC (Conversational Sarcasm Corpus; \citealp{jang2024CSC}) datasets have similarly low to moderate agreements (0.23--0.59) for scripted language from TV shows and natural sarcasm utterances respectively, which underscores the annotation ambiguity of indirect language that also applies to IQA.
Additionally, the perspectivism research of \citet{casola-etal-2024-multipico} reveals that for irony detection, the results vary greatly between different demographic groups. It is important to track the annotators' backgrounds and meta information and take it into account for analyses, as the subjectiveness of different groups may influence the annotations.

\section{Conclusions}\label{sec:conclusions}
We presented InQA+, a multilingual evaluation dataset, and GenIQA, an LLM-generated training corpus for IQA in English, German, and Bavarian. 
Our experiments confirm that IQA is pragmatically hard for high- and low-resource languages and we find that a large dataset is beneficial for good performance. Nevertheless, data quality and annotation/label clarity are decisive factors. Simplified labels help mitigate ambiguities, but come at the cost of loss of informational nuances.

Therefore, we suggest conducting intensive pre-experiments to check that the training dataset leads to models of sufficient quality that do not overfit. Additionally, we recommend to invest in resources and ways to create sufficiently big datasets (also for languages other than English), as this is crucial for the performance. 

We will release our new datasets, \inqaplus\ and \geniqa, to enable further research on indirect question answering in high- and low-resource languages and on manually vs.\ automatically generated datasets.

\section*{Limitations}
We experiment with a limited selection of two high-resource languages and one low-resource dialect, because IQA resources are sparse as \S\ref{sec:related_work} shows. 
For future work, it is possible to repeat the data acquisition method with the other languages in the opensubtitles \cite{lison2016opensubtitles2018} corpus to create more multilingual IQA corpora.

The data we used stems from movie subtitles provided by opensubtitles v2018 \cite{lison2016opensubtitles2018} and contains scripted language that is not necessarily naturally occurring. This means that the performance of real-world data might lead to different results.

\section*{Ethical Considerations}
The Bavarian translation and annotation of the data was performed by an employed worker who is paid according to national standards. The data does not contain any personally identifiable information and is used according to the permissions of the dataset provider.

\section*{Acknowledgements}
We thank the anonymous reviewers as well as the members of the MaiNLP research lab for their constructive feedback, especially Rob van der Goot and Felicia Körner.

The work presented in this paper was funded by ERC Consolidator Grant DIALECT 101043235.

\vspace{0.5cm}

\nocite{*}
\section{Bibliographical References}\label{sec:reference}
\bibliographystyle{lrec2026-natbib}
\bibliography{lrec2026-example}

\begin{thebibliography}{66}
\expandafter\ifx\csname natexlab\endcsname\relax\def\natexlab#1{#1}\fi

\bibitem[{Awwalu et~al.(2020)Awwalu, Abdullahi, and Evwiekpaefe}]{awwalu2020pos}
Jamilu Awwalu, Saleh El-Yakub Abdullahi, and Abraham~Eseoghene Evwiekpaefe. 2020.
\newblock \href {https://doi.org/https://doi.org/10.33003/fjs-2020-0402-325} {Parts of speech tagging: A review of techniques}.
\newblock \emph{FUDMA JOURNAL OF SCIENCES}, 4(2):712--721.

\bibitem[{Blaschke et~al.(2024{\natexlab{a}})Blaschke, Kova{\v{c}}i{\'c}, Peng, and Plank}]{blaschke2024maibaamguidelines}
Verena Blaschke, Barbara Kova{\v{c}}i{\'c}, Siyao Peng, and Barbara Plank. 2024{\natexlab{a}}.
\newblock Maibaam annotation guidelines.
\newblock \emph{arXiv preprint arXiv:2403.05902}.

\bibitem[{Blaschke et~al.(2024{\natexlab{b}})Blaschke, Kova{\v{c}}i{\'c}, Peng, Sch{\"u}tze, and Plank}]{blaschke2024maibaam}
Verena Blaschke, Barbara Kova{\v{c}}i{\'c}, Siyao Peng, Hinrich Sch{\"u}tze, and Barbara Plank. 2024{\natexlab{b}}.
\newblock \href {https://aclanthology.org/2024.lrec-main.953/} {{M}ai{B}aam: A multi-dialectal {B}avarian {U}niversal {D}ependency treebank}.
\newblock In \emph{Proceedings of the 2024 Joint International Conference on Computational Linguistics, Language Resources and Evaluation (LREC-COLING 2024)}, pages 10921--10938, Torino, Italia. ELRA and ICCL.

\bibitem[{Blaschke et~al.(2024{\natexlab{c}})Blaschke, Purschke, Schuetze, and Plank}]{blaschke2024dialectspeakers}
Verena Blaschke, Christoph Purschke, Hinrich Schuetze, and Barbara Plank. 2024{\natexlab{c}}.
\newblock \href {https://doi.org/10.18653/v1/2024.acl-short.74} {What do dialect speakers want? a survey of attitudes towards language technology for {G}erman dialects}.
\newblock In \emph{Proceedings of the 62nd Annual Meeting of the Association for Computational Linguistics (Volume 2: Short Papers)}, pages 823--841, Bangkok, Thailand. Association for Computational Linguistics.

\bibitem[{Blaschke et~al.(2026)Blaschke, Winkler, and Plank}]{blaschke-etal-2026-standard}
Verena Blaschke, Miriam Winkler, and Barbara Plank. 2026.
\newblock \href {http://arxiv.org/abs/2510.07890} {Standard-to-dialect transfer trends differ across text and speech: A case study on intent and topic classification in {German} dialects}.

\bibitem[{Bowman et~al.(2015)Bowman, Angeli, Potts, and Manning}]{bowman2015nli}
Samuel~R. Bowman, Gabor Angeli, Christopher Potts, and Christopher~D. Manning. 2015.
\newblock \href {https://doi.org/10.18653/v1/D15-1075} {A large annotated corpus for learning natural language inference}.
\newblock In \emph{Proceedings of the 2015 Conference on Empirical Methods in Natural Language Processing}, pages 632--642, Lisbon, Portugal. Association for Computational Linguistics.

\bibitem[{{CAIDAS Uni Würzburg}(2024)}]{caidas2025betzerl}
{CAIDAS Uni Würzburg}. 2024.
\newblock Betzerl.
\newblock \url{https://huggingface.co/LSX-UniWue/Betzerl_1B_wiki_preview}.

\bibitem[{Casola et~al.(2024)Casola, Frenda, Lo, Sezerer, Uva, Basile, Bosco, Pedrani, Rubagotti, Patti, and Bernardi}]{casola-etal-2024-multipico}
Silvia Casola, Simona Frenda, Soda~Marem Lo, Erhan Sezerer, Antonio Uva, Valerio Basile, Cristina Bosco, Alessandro Pedrani, Chiara Rubagotti, Viviana Patti, and Davide Bernardi. 2024.
\newblock \href {https://doi.org/10.18653/v1/2024.acl-long.849} {{M}ulti{PIC}o: Multilingual perspectivist irony corpus}.
\newblock In \emph{Proceedings of the 62nd Annual Meeting of the Association for Computational Linguistics (Volume 1: Long Papers)}, pages 16008--16021, Bangkok, Thailand. Association for Computational Linguistics.

\bibitem[{Castro et~al.(2019)Castro, Hazarika, Pérez-Rosas, Zimmermann, Mihalcea, and Poria}]{castro2019mustard}
Santiago Castro, Devamanyu Hazarika, Verónica Pérez-Rosas, Roger Zimmermann, Rada Mihalcea, and Soujanya Poria. 2019.
\newblock \href {http://arxiv.org/abs/1906.01815} {Towards multimodal sarcasm detection (an \emph{obviously} perfect paper)}.

\bibitem[{Chung et~al.(2022)Chung, Hou, Longpre, Zoph, Tay, Fedus, Li, Wang, Dehghani, Brahma, Webson, Gu, Dai, Suzgun, Chen, Chowdhery, Narang, Mishra, Yu, Zhao, Huang, Dai, Yu, Petrov, Chi, Dean, Devlin, Roberts, Zhou, Le, and Wei}]{chung2022flant5xxl}
Hyung~Won Chung, Le~Hou, Shayne Longpre, Barret Zoph, Yi~Tay, William Fedus, Eric Li, Xuezhi Wang, Mostafa Dehghani, Siddhartha Brahma, Albert Webson, Shixiang~Shane Gu, Zhuyun Dai, Mirac Suzgun, Xinyun Chen, Aakanksha Chowdhery, Sharan Narang, Gaurav Mishra, Adams Yu, Vincent Zhao, Yanping Huang, Andrew Dai, Hongkun Yu, Slav Petrov, Ed~H. Chi, Jeff Dean, Jacob Devlin, Adam Roberts, Denny Zhou, Quoc~V. Le, and Jason Wei. 2022.
\newblock \href {https://doi.org/10.48550/ARXIV.2210.11416} {Scaling instruction-finetuned language models}.

\bibitem[{Conneau et~al.(2020)Conneau, Khandelwal, Goyal, Chaudhary, Wenzek, Guzm{\'a}n, Grave, Ott, Zettlemoyer, and Stoyanov}]{conneau2020xmlroberta}
Alexis Conneau, Kartikay Khandelwal, Naman Goyal, Vishrav Chaudhary, Guillaume Wenzek, Francisco Guzm{\'a}n, Edouard Grave, Myle Ott, Luke Zettlemoyer, and Veselin Stoyanov. 2020.
\newblock \href {https://doi.org/10.18653/v1/2020.acl-main.747} {Unsupervised cross-lingual representation learning at scale}.
\newblock In \emph{Proceedings of the 58th Annual Meeting of the Association for Computational Linguistics}, pages 8440--8451, Online. Association for Computational Linguistics.

\bibitem[{Damgaard et~al.(2021)Damgaard, Toborek, Eriksen, and Plank}]{damgaard2021friendsqia}
Cathrine Damgaard, Paulina Toborek, Trine Eriksen, and Barbara Plank. 2021.
\newblock \href {https://doi.org/10.18653/v1/2021.codi-main.1} {{\textquotedblleft}{I}`ll be there for you{\textquotedblright}: The one with understanding indirect answers}.
\newblock In \emph{Proceedings of the 2nd Workshop on Computational Approaches to Discourse}, pages 1--11, Punta Cana, Dominican Republic and Online. Association for Computational Linguistics.

\bibitem[{Deore et~al.(2025)Deore, Patil, and Patil}]{deore2025poslowresource}
Pallavi~R. Deore, Nita~V. Patil, and Ajay~S. Patil. 2025.
\newblock \href {https://doi.org/https://doi.org/10.1016/j.procs.2025.04.632} {Deep learning-based parts-of-speech tagging in marathi language}.
\newblock \emph{Procedia Computer Science}, 258:3771--3780.
\newblock International Conference on Machine Learning and Data Engineering.

\bibitem[{Devlin et~al.(2019)Devlin, Chang, Lee, and Toutanova}]{devlin2019bert}
Jacob Devlin, Ming-Wei Chang, Kenton Lee, and Kristina Toutanova. 2019.
\newblock \href {https://doi.org/10.18653/v1/N19-1423} {{BERT}: Pre-training of deep bidirectional transformers for language understanding}.
\newblock In \emph{Proceedings of the 2019 Conference of the North {A}merican Chapter of the Association for Computational Linguistics: Human Language Technologies, Volume 1 (Long and Short Papers)}, pages 4171--4186, Minneapolis, Minnesota. Association for Computational Linguistics.

\bibitem[{Dubey et~al.(2025)Dubey, Dubey, and Bokoro}]{dubey2025sarcasm}
Parul Dubey, Pushkar Dubey, and Pitshou~N. Bokoro. 2025.
\newblock \href {https://doi.org/10.3390/computers14030095} {Unpacking sarcasm: A contextual and transformer-based approach for improved detection}.
\newblock \emph{Computers}, 14(3).

\bibitem[{Dušková(1981)}]{duskova1981negativequestions}
Libuše Dušková. 1981.
\newblock \href {https://doi.org/doi:10.1515/iral.1981.19.1-4.181} {Negative questions in english}.
\newblock \emph{International Review of Applied Linguistics in Language Teaching}, 19(1-4):181--194.

\bibitem[{{Gemma Team}(2024)}]{gemmateam2024gemma2}
{Gemma Team}. 2024.
\newblock \href {http://arxiv.org/abs/2408.00118} {Gemma 2: Improving open language models at a practical size}.

\bibitem[{Glenwright et~al.(2014)Glenwright, Parackel, Cheung, and Nilsen}]{glenwright2014sarcasmnintonation}
Melanie Glenwright, Jayanthi~M. Parackel, Kristene R.~J. Cheung, and Elizabeth~S. Nilsen. 2014.
\newblock \href {https://doi.org/10.1017/S0305000912000773} {Intonation influences how children and adults interpret sarcasm}.
\newblock \emph{Journal of Child Language}, 41(2):472–484.

\bibitem[{Haq et~al.(2025)Haq, Zhang, and Qadri}]{haq2025pos}
Ijazul Haq, Yingjie Zhang, and Intakhab~Alam Qadri. 2025.
\newblock \href {https://doi.org/10.1007/s10579-025-09834-3} {Pos tagging of low-resource pashto language: annotated corpus and bert-based model}.
\newblock \emph{Lang Resources \& Evaluation 59}, pages 3243–--3265.

\bibitem[{He et~al.(2020)He, Liu, Gao, and Chen}]{he2020mdeberta}
Pengcheng He, Xiaodong Liu, Jianfeng Gao, and Weizhu Chen. 2020.
\newblock \href {http://arxiv.org/abs/2006.03654} {Deberta: Decoding-enhanced {BERT} with disentangled attention}.
\newblock \emph{CoRR}, abs/2006.03654.

\bibitem[{Holmberg(2012)}]{holmberg2012syntax}
Anders Holmberg. 2012.
\newblock The syntax of negative questions and their answers.
\newblock \emph{Proceedings of GLOW in Asia IX}.

\bibitem[{Holmberg(2013)}]{holmberg2013syntax}
Anders Holmberg. 2013.
\newblock The syntax of answers to polar questions in english and swedish.
\newblock \emph{Lingua}, 128:31--50.

\bibitem[{Hu et~al.(2023)Hu, Floyd, Jouravlev, Fedorenko, and Gibson}]{hu2023finegrainedcomparisonpragmaticlanguage}
Jennifer Hu, Sammy Floyd, Olessia Jouravlev, Evelina Fedorenko, and Edward Gibson. 2023.
\newblock \href {http://arxiv.org/abs/2212.06801} {A fine-grained comparison of pragmatic language understanding in humans and language models}.

\bibitem[{Jang and Frassinelli(2024)}]{jang2024CSC}
Hyewon Jang and Diego Frassinelli. 2024.
\newblock \href {https://doi.org/10.18653/v1/2024.naacl-long.238} {Generalizable sarcasm detection is just around the corner, of course!}
\newblock In \emph{Proceedings of the 2024 Conference of the North American Chapter of the Association for Computational Linguistics: Human Language Technologies (Volume 1: Long Papers)}, pages 4238--4249, Mexico City, Mexico. Association for Computational Linguistics.

\bibitem[{Jayaraman et~al.(2022)Jayaraman, Trueman, Ananthakrishnan, Mitra, Liu, and Cambria}]{jayaraman2022sarcasm}
Ashok~Kumar Jayaraman, Tina~Esther Trueman, Gayathri Ananthakrishnan, Satanik Mitra, Qian Liu, and Erik Cambria. 2022.
\newblock \href {https://doi.org/10.1109/AIDE57180.2022.10060855} {Sarcasm detection in news headlines using supervised learning}.
\newblock In \emph{2022 International Conference on Artificial Intelligence and Data Engineering (AIDE)}, pages 288--294.

\bibitem[{Khan et~al.(2025)Khan, Qasim, Khan, Aurangzeb, Khan, and Anwar}]{khan2025sarcasmlowresource}
Shumaila Khan, Iqbal Qasim, Wahab Khan, Khursheed Aurangzeb, Javed~Ali Khan, and Muhammad~Shahid Anwar. 2025.
\newblock \href {https://doi.org/https://doi.org/10.1111/exsy.13686} {A novel transformer attention-based approach for sarcasm detection}.
\newblock \emph{Expert Systems}, 42(1):e13686.

\bibitem[{Kr{\"u}ckl et~al.(2025)Kr{\"u}ckl, Blaschke, and Plank}]{krueckl2025xsid}
Xaver~Maria Kr{\"u}ckl, Verena Blaschke, and Barbara Plank. 2025.
\newblock \href {https://aclanthology.org/2025.vardial-1.10/} {Improving dialectal slot and intent detection with auxiliary tasks: A multi-dialectal {B}avarian case study}.
\newblock In \emph{Proceedings of the 12th Workshop on NLP for Similar Languages, Varieties and Dialects}, pages 128--146, Abu Dhabi, UAE. Association for Computational Linguistics.

\bibitem[{Krückl et~al.(2025)Krückl, Blaschke, and Plank}]{krückl2025sid}
Xaver~Maria Krückl, Verena Blaschke, and Barbara Plank. 2025.
\newblock \href {http://arxiv.org/abs/2501.03863} {Improving dialectal slot and intent detection with auxiliary tasks: A multi-dialectal bavarian case study}.

\bibitem[{Kuno(1975)}]{kuno1975structure}
Susumu Kuno. 1975.
\newblock The structure of the japanese language.
\newblock \emph{Foundations of Language}, 13(3).

\bibitem[{Leiner(2025)}]{leiner2025sosci}
Dominik~J. Leiner. 2025.
\newblock Sosci survey (version 3.7.06).
\newblock \url{https://www.soscisurvey.de}.

\bibitem[{Li et~al.(2022)Li, Mao, and Wang}]{li2022pos}
Hongwei Li, Hongyan Mao, and Jingzi Wang. 2022.
\newblock \href {https://doi.org/10.3390/electronics11010056} {Part-of-speech tagging with rule-based data preprocessing and transformer}.
\newblock \emph{Electronics}, 11(1).

\bibitem[{Lison and Tiedemann(2016)}]{lison2016opensubtitles2018}
Pierre Lison and J{\"o}rg Tiedemann. 2016.
\newblock \href {https://aclanthology.org/L16-1147/} {{O}pen{S}ubtitles2016: Extracting large parallel corpora from movie and {TV} subtitles}.
\newblock In \emph{Proceedings of the Tenth International Conference on Language Resources and Evaluation ({LREC}`16)}, pages 923--929, Portoro{\v{z}}, Slovenia. European Language Resources Association (ELRA).

\bibitem[{Louis et~al.(2020)Louis, Roth, and Radlinski}]{louis2020circa}
Annie Louis, Dan Roth, and Filip Radlinski. 2020.
\newblock \href {https://doi.org/10.18653/v1/2020.emnlp-main.601} {{\textquotedblleft}{I}`d rather just go to bed{\textquotedblright}: Understanding indirect answers}.
\newblock In \emph{Proceedings of the 2020 Conference on Empirical Methods in Natural Language Processing (EMNLP)}, pages 7411--7425, Online. Association for Computational Linguistics.

\bibitem[{Ma et~al.(2025)Ma, Li, Zhou, Gong, Liu, Jasinskaja, Friedrich, Hirschberg, Kreuter, and Plank}]{ma2025pragmaticseralargelanguage}
Bolei Ma, Yuting Li, Wei Zhou, Ziwei Gong, Yang~Janet Liu, Katja Jasinskaja, Annemarie Friedrich, Julia Hirschberg, Frauke Kreuter, and Barbara Plank. 2025.
\newblock \href {http://arxiv.org/abs/2502.12378} {Pragmatics in the era of large language models: A survey on datasets, evaluation, opportunities and challenges}.

\bibitem[{Martins et~al.(2025)Martins, Fernandes, Alves, Guerreiro, Rei, Alves, Pombal, Farajian, Faysse, Klimaszewski, Colombo, Haddow, {de Souza}, Birch, and Martins}]{martins2025eurollm}
Pedro~Henrique Martins, Patrick Fernandes, João Alves, Nuno~M. Guerreiro, Ricardo Rei, Duarte~M. Alves, José Pombal, Amin Farajian, Manuel Faysse, Mateusz Klimaszewski, Pierre Colombo, Barry Haddow, José~G.C. {de Souza}, Alexandra Birch, and André~F.T. Martins. 2025.
\newblock \href {https://doi.org/https://doi.org/10.1016/j.procs.2025.02.260} {Eurollm: Multilingual language models for europe}.
\newblock \emph{Procedia Computer Science}, 255:53--62.
\newblock Proceedings of the Second EuroHPC user day.

\bibitem[{Meta(2024)}]{meta2024llama}
Meta. 2024.
\newblock Llama 3.2: Revolutionizing edge ai and vision with open, customizable models.
\newblock https://ai.meta.com/blog/llama-3-2-connect-2024-vision-edge-mobile-devices/.

\bibitem[{Mitri et~al.(2025)Mitri, Lyngdoh, Warjri, Saha, Lyngdoh, and Maji}]{Mitri2025POS}
Aiom~Minnette Mitri, Eusebius~Lawai Lyngdoh, Sunita Warjri, Goutam Saha, Saralin~A. Lyngdoh, and Arnab~Kumar Maji. 2025.
\newblock \href {https://doi.org/10.1017/nlp.2024.24} {Probing a pretrained roberta on khasi language for pos tagging}.
\newblock \emph{Natural Language Processing}, 31(2):230–249.

\bibitem[{Mohammad et~al.(2025)Mohammad, Abdullahi, and Achir}]{mohammad2025pos}
Abdukarim Mohammad, Mohammed Abdullahi, and Jerome~Aondongu Achir. 2025.
\newblock \href {https://doi.org/10.56705/ijodas.v6i1.184} {Improving part-of-speech tagging with relative positional encoding in transformer models and basic rules}.
\newblock \emph{Indonesian Journal of Data and Science}, 6(1):10--19.

\bibitem[{Mosbach et~al.(2023)Mosbach, Pimentel, Ravfogel, Klakow, and Elazar}]{mosbach-etal-2023-shot}
Marius Mosbach, Tiago Pimentel, Shauli Ravfogel, Dietrich Klakow, and Yanai Elazar. 2023.
\newblock \href {https://doi.org/10.18653/v1/2023.findings-acl.779} {Few-shot fine-tuning vs. in-context learning: A fair comparison and evaluation}.
\newblock In \emph{Findings of the Association for Computational Linguistics: ACL 2023}, pages 12284--12314, Toronto, Canada. Association for Computational Linguistics.

\bibitem[{M{\"u}ller and Plank(2024)}]{mueller2024indirectqa}
Christin M{\"u}ller and Barbara Plank. 2024.
\newblock \href {https://aclanthology.org/2024.lrec-main.791/} {{I}ndirect{QA}: Understanding indirect answers to implicit polar questions in {F}rench and {S}panish}.
\newblock In \emph{Proceedings of the 2024 Joint International Conference on Computational Linguistics, Language Resources and Evaluation (LREC-COLING 2024)}, pages 9025--9035, Torino, Italia. ELRA and ICCL.

\bibitem[{Naseem et~al.(2009)Naseem, Snyder, Eisenstein, and Barzilay}]{naseem2009multilingualpos}
Tahira Naseem, Benjamin Snyder, Jacob Eisenstein, and Regina Barzilay. 2009.
\newblock Multilingual part-of-speech tagging: Two unsupervised approaches.
\newblock \emph{Journal of Artificial Intelligence Research}, 36:341--385.

\bibitem[{Ning et~al.(2025)Ning, Zhang, Ye, Guo, and Guan}]{ning2025nli}
Meiling Ning, Zhongbao Zhang, Junda Ye, Jiabao Guo, and Qingyuan Guan. 2025.
\newblock \href {http://arxiv.org/abs/2508.18212} {Better language model-based judging reward modeling through scaling comprehension boundaries}.

\bibitem[{OLMo et~al.(2024)OLMo, Walsh, Soldaini, Groeneveld, Lo, Arora, Bhagia, Gu, Huang, Jordan, Lambert, Schwenk, Tafjord, Anderson, Atkinson, Brahman, Clark, Dasigi, Dziri, Guerquin, Ivison, Koh, Liu, Malik, Merrill, Miranda, Morrison, Murray, Nam, Pyatkin, Rangapur, Schmitz, Skjonsberg, Wadden, Wilhelm, Wilson, Zettlemoyer, Farhadi, Smith, and Hajishirzi}]{olmo20242olmo2furious}
Team OLMo, Pete Walsh, Luca Soldaini, Dirk Groeneveld, Kyle Lo, Shane Arora, Akshita Bhagia, Yuling Gu, Shengyi Huang, Matt Jordan, Nathan Lambert, Dustin Schwenk, Oyvind Tafjord, Taira Anderson, David Atkinson, Faeze Brahman, Christopher Clark, Pradeep Dasigi, Nouha Dziri, Michal Guerquin, Hamish Ivison, Pang~Wei Koh, Jiacheng Liu, Saumya Malik, William Merrill, Lester James~V. Miranda, Jacob Morrison, Tyler Murray, Crystal Nam, Valentina Pyatkin, Aman Rangapur, Michael Schmitz, Sam Skjonsberg, David Wadden, Christopher Wilhelm, Michael Wilson, Luke Zettlemoyer, Ali Farhadi, Noah~A. Smith, and Hannaneh Hajishirzi. 2024.
\newblock \href {http://arxiv.org/abs/2501.00656} {2 olmo 2 furious}.

\bibitem[{OpenAI(2024)}]{openai2024gpt}
OpenAI. 2024.
\newblock \href {https://platform.openai.com/docs} {Openai api documentation}.

\bibitem[{Oraby et~al.(2017)Oraby, Harrison, Reed, Hernandez, Riloff, and Walker}]{oraby2017sarcasmcorpus}
Shereen Oraby, Vrindavan Harrison, Lena Reed, Ernesto Hernandez, Ellen Riloff, and Marilyn Walker. 2017.
\newblock \href {http://arxiv.org/abs/1709.05404} {Creating and characterizing a diverse corpus of sarcasm in dialogue}.

\bibitem[{Paci et~al.(2025)Paci, Panunzi, and Pezzelle}]{paci2025llmimplicity}
Walter Paci, Alessandro Panunzi, and Sandro Pezzelle. 2025.
\newblock \href {http://arxiv.org/abs/2506.06775} {They want to pretend not to understand: The limits of current llms in interpreting implicit content of political discourse}.

\bibitem[{Peng et~al.(2024)Peng, Sun, Shan, Kolm, Blaschke, Artemova, and Plank}]{peng2024barner}
Siyao Peng, Zihang Sun, Huangyan Shan, Marie Kolm, Verena Blaschke, Ekaterina Artemova, and Barbara Plank. 2024.
\newblock \href {https://aclanthology.org/2024.lrec-main.1262/} {{S}ebastian, {B}asti, {W}astl?! recognizing named entities in {B}avarian dialectal data}.
\newblock In \emph{Proceedings of the 2024 Joint International Conference on Computational Linguistics, Language Resources and Evaluation (LREC-COLING 2024)}, pages 14478--14493, Torino, Italia. ELRA and ICCL.

\bibitem[{Pfister et~al.(2024)Pfister, Wunderle, and Hotho}]{pfister2024llammlein}
Jan Pfister, Julia Wunderle, and Andreas Hotho. 2024.
\newblock \href {http://arxiv.org/abs/2411.11171} {Ll\"ammlein: Compact and competitive german-only language models from scratch}.

\bibitem[{Plank(2022)}]{plank-2022-problem}
Barbara Plank. 2022.
\newblock \href {https://doi.org/10.18653/v1/2022.emnlp-main.731} {The ``problem'' of human label variation: On ground truth in data, modeling and evaluation}.
\newblock In \emph{Proceedings of the 2022 Conference on Empirical Methods in Natural Language Processing}, pages 10671--10682, Abu Dhabi, United Arab Emirates. Association for Computational Linguistics.

\bibitem[{Plank et~al.(2016)Plank, S{\o}gaard, and Goldberg}]{plank2016pos}
Barbara Plank, Anders S{\o}gaard, and Yoav Goldberg. 2016.
\newblock \href {https://doi.org/10.18653/v1/P16-2067} {Multilingual part-of-speech tagging with bidirectional long short-term memory models and auxiliary loss}.
\newblock In \emph{Proceedings of the 54th Annual Meeting of the Association for Computational Linguistics (Volume 2: Short Papers)}, pages 412--418, Berlin, Germany. Association for Computational Linguistics.

\bibitem[{Qiu et~al.(2023)Qiu, Duan, and Cai}]{qiu-etal-2023-chatgpt}
Zhuang Qiu, Xufeng Duan, and Zhenguang Cai. 2023.
\newblock \href {https://aclanthology.org/2023.naloma-1.3/} {Does {C}hat{GPT} resemble humans in processing implicatures?}
\newblock In \emph{Proceedings of the 4th Natural Logic Meets Machine Learning Workshop}, pages 25--34, Nancy, France. Association for Computational Linguistics.

\bibitem[{Qorib et~al.(2024)Qorib, Moon, and Ng}]{qorib-etal-2024-decoder}
Muhammad~Reza Qorib, Geonsik Moon, and Hwee~Tou Ng. 2024.
\newblock \href {https://doi.org/10.18653/v1/2024.findings-acl.967} {Are decoder-only language models better than encoder-only language models in understanding word meaning?}
\newblock In \emph{Findings of the Association for Computational Linguistics: ACL 2024}, pages 16339--16347, Bangkok, Thailand. Association for Computational Linguistics.

\bibitem[{Sanagavarapu et~al.(2022)Sanagavarapu, Singaraju, Kakileti, Kaza, Mathews, Li, Brito, and Blanco}]{sanagavarapu2022swdaia}
Krishna Sanagavarapu, Jathin Singaraju, Anusha Kakileti, Anirudh Kaza, Aaron Mathews, Helen Li, Nathan Brito, and Eduardo Blanco. 2022.
\newblock \href {https://doi.org/10.18653/v1/2022.naacl-main.345} {Disentangling indirect answers to yes-no questions in real conversations}.
\newblock In \emph{Proceedings of the 2022 Conference of the North American Chapter of the Association for Computational Linguistics: Human Language Technologies}, pages 4677--4695, Seattle, United States. Association for Computational Linguistics.

\bibitem[{Snyder et~al.(2008)Snyder, Naseem, Eisenstein, and Barzilay}]{snyder2008multilingualpos}
Benjamin Snyder, Tahira Naseem, Jacob Eisenstein, and Regina Barzilay. 2008.
\newblock \href {https://aclanthology.org/D08-1109/} {Unsupervised multilingual learning for {POS} tagging}.
\newblock In \emph{Proceedings of the 2008 Conference on Empirical Methods in Natural Language Processing}, pages 1041--1050, Honolulu, Hawaii. Association for Computational Linguistics.

\bibitem[{Sravanthi et~al.(2024)Sravanthi, Doshi, Tankala, Murthy, Dabre, and Bhattacharyya}]{sravanthi2024pub}
Settaluri Sravanthi, Meet Doshi, Pavan Tankala, Rudra Murthy, Raj Dabre, and Pushpak Bhattacharyya. 2024.
\newblock \href {https://doi.org/10.18653/v1/2024.findings-acl.719} {{PUB}: A pragmatics understanding benchmark for assessing {LLM}s' pragmatics capabilities}.
\newblock In \emph{Findings of the Association for Computational Linguistics: ACL 2024}, pages 12075--12097, Bangkok, Thailand. Association for Computational Linguistics.

\bibitem[{Storks et~al.(2020)Storks, Gao, and Chai}]{storks2020nli}
Shane Storks, Qiaozi Gao, and Joyce~Y. Chai. 2020.
\newblock \href {http://arxiv.org/abs/1904.01172} {Recent advances in natural language inference: A survey of benchmarks, resources, and approaches}.

\bibitem[{Sujon et~al.(2025)Sujon, Hassan, Choi, and Samad}]{sujon2025f1score}
Khaled~Mahmud Sujon, Rohayanti Hassan, Kwonhue Choi, and Md~Abdus Samad. 2025.
\newblock Accuracy, precision, recall, f1-score, or {MCC}? empirical evidence from advanced statistics, {ML}, and {XAI} for evaluating business predictive models.
\newblock \emph{Journal of Big Data}, 12(268).

\bibitem[{Ulmer et~al.(2022)Ulmer, Bassignana, M{\"u}ller-Eberstein, Varab, Zhang, van~der Goot, Hardmeier, and Plank}]{ulmer2022experimentalstandards}
Dennis Ulmer, Elisa Bassignana, Max M{\"u}ller-Eberstein, Daniel Varab, Mike Zhang, Rob van~der Goot, Christian Hardmeier, and Barbara Plank. 2022.
\newblock \href {https://doi.org/10.18653/v1/2022.findings-emnlp.196} {Experimental standards for deep learning in natural language processing research}.
\newblock In \emph{Findings of the Association for Computational Linguistics: EMNLP 2022}, pages 2673--2692, Abu Dhabi, United Arab Emirates. Association for Computational Linguistics.

\bibitem[{van~der Goot et~al.(2021)van~der Goot, Sharaf, Imankulova, {\"U}st{\"u}n, Stepanovi{\'c}, Ramponi, Khairunnisa, Komachi, and Plank}]{vandergoot2021xsid}
Rob van~der Goot, Ibrahim Sharaf, Aizhan Imankulova, Ahmet {\"U}st{\"u}n, Marija Stepanovi{\'c}, Alan Ramponi, Siti~Oryza Khairunnisa, Mamoru Komachi, and Barbara Plank. 2021.
\newblock \href {https://doi.org/10.18653/v1/2021.naacl-main.197} {From masked language modeling to translation: Non-{E}nglish auxiliary tasks improve zero-shot spoken language understanding}.
\newblock In \emph{Proceedings of the 2021 Conference of the North American Chapter of the Association for Computational Linguistics: Human Language Technologies}, pages 2479--2497, Online. Association for Computational Linguistics.

\bibitem[{Vaswani et~al.(2017)Vaswani, Shazeer, Parmar, Uszkoreit, Jones, Gomez, Kaiser, and Polosukhin}]{vaswani2017transformers}
Ashish Vaswani, Noam Shazeer, Niki Parmar, Jakob Uszkoreit, Llion Jones, Aidan~N Gomez, \L~ukasz Kaiser, and Illia Polosukhin. 2017.
\newblock \href {https://proceedings.neurips.cc/paper_files/paper/2017/file/3f5ee243547dee91fbd053c1c4a845aa-Paper.pdf} {Attention is all you need}.
\newblock In \emph{Advances in Neural Information Processing Systems}, volume~30. Curran Associates, Inc.

\bibitem[{Wang et~al.(2023)Wang, Hossain, Mathur, Melo, Ozler, Park, Quintero, Rezaei, Shakya, Uddin, and Blanco}]{wang2023indirectanswers}
Zijie Wang, Md~Hossain, Shivam Mathur, Terry Melo, Kadir Ozler, Keun Park, Jacob Quintero, MohammadHossein Rezaei, Shreya Shakya, Md~Uddin, and Eduardo Blanco. 2023.
\newblock \href {https://doi.org/10.18653/v1/2023.findings-emnlp.146} {Interpreting indirect answers to yes-no questions in multiple languages}.
\newblock In \emph{Findings of the Association for Computational Linguistics: EMNLP 2023}, pages 2210--2227, Singapore. Association for Computational Linguistics.

\bibitem[{Wei{\ss}(1998)}]{weiss1998bavariangrammar}
Helmut Wei{\ss}. 1998.
\newblock \emph{Syntax des Bairischen: Studien zur Grammatik einer natürlichen Sprache}.
\newblock Max Niemeyer Verlag.

\bibitem[{Weller et~al.(2025)Weller, Ricci, Marone, Chaffin, Lawrie, and Durme}]{weller2025seqvsseqopen}
Orion Weller, Kathryn Ricci, Marc Marone, Antoine Chaffin, Dawn Lawrie, and Benjamin~Van Durme. 2025.
\newblock \href {http://arxiv.org/abs/2507.11412} {Seq vs seq: An open suite of paired encoders and decoders}.

\bibitem[{Winkler et~al.(2024)Winkler, Juozapaityte, van~der Goot, and Plank}]{winkler2024sid}
Miriam Winkler, Virginija Juozapaityte, Rob van~der Goot, and Barbara Plank. 2024.
\newblock \href {https://aclanthology.org/2024.lrec-main.1297/} {Slot and intent detection resources for {B}avarian and {L}ithuanian: Assessing translations vs natural queries to digital assistants}.
\newblock In \emph{Proceedings of the 2024 Joint International Conference on Computational Linguistics, Language Resources and Evaluation (LREC-COLING 2024)}, pages 14898--14915, Torino, Italia. ELRA and ICCL.

\bibitem[{Zhang et~al.(2016)Zhang, Bengio, Hardt, Recht, and Vinyals}]{zhang2016understanding}
Chiyuan Zhang, Samy Bengio, Moritz Hardt, Benjamin Recht, and Oriol Vinyals. 2016.
\newblock Understanding deep learning requires rethinking generalization.
\newblock \emph{arXiv preprint arXiv:1611.03530}.

\bibitem[{Zhang et~al.(2021)Zhang, Liu, Li, Tiwari, Wang, Li, Pandey, Zhang, and Song}]{zhang2021sarcasm}
Yazhou Zhang, Yaochen Liu, Qiuchi Li, Prayag Tiwari, Benyou Wang, Yuhua Li, Hari~Mohan Pandey, Peng Zhang, and Dawei Song. 2021.
\newblock \href {https://doi.org/10.1109/TFUZZ.2021.3072492} {Cfn: A complex-valued fuzzy network for sarcasm detection in conversations}.
\newblock \emph{IEEE Transactions on Fuzzy Systems}, 29(12):3696--3710.

\end{thebibliography}

\appendix

\section{Data Statements}\label{app:data_statement}
The information in \S\ref{sec:appendix-data-distribution} and~\S\ref{sec:appendix-data-acknowledgements} applies to both \inqaplus~(\S\ref{app:inqaplus_statement}) and\geniqa~(\S\ref{app:geniqa_statement}).

\subsection{\inqaplus\ Data Statement}\label{app:inqaplus_statement}
\begin{itemize}
    \item Dataset name and version: \inqaplus\ (EN, DE and BAR) (version 0.1)
    \item Dataset curators: Miriam Winkler, Verena Blaschke, Barbara Plank
    \item Dataset and data statement citation: Please cite this paper.
    \item Data statement version: 1.0
\end{itemize}

\noindent
\textbf{Executive summary and curation rationale.}
This dataset extends the work of \citet{mueller2024indirectqa} and adds a new Indirect Question Answering (IQA) resource. It contains 438 indirect question-answers per language in English, Standard German and Bavarian, a German dialect. It serves as an evaluation dataset for IQA.

\textbf{Documentation for source datasets.}
\label{sec:datasheet-source-datasets}
We use data from the opensubtitles v2018 corpus \cite{lison2016opensubtitles2018}. The subtitles stem from the website \url{http://www.opensubtitles.org/}, which made the data available for research purposes. The subtitles are aligned between different languages. 

The dataset consists of questions and indirect answers from movie scripts with numeric labels indicating the polarity of the indirect answers. They are separated by tabspaces with one question-answer pair and label per line.

\textbf{Date.} 
The movies stem from different years and are identifiable through the meta data provided in the opensubtitles v2018 corpus \cite{lison2016opensubtitles2018}. The movies' release dates range from the 19th to 21st century. 

\textbf{Modality.} Written (movie subtitles).

\textbf{Genre.} Questions and indirect answers from movie scripts.

\textbf{Dataset statistics.}
The dataset consists of 438 English sentences and a German and Bavarian translation for each. It consists only of a test split for each language.

\textbf{Labels.}
Each question-answer pair carries a numerical label indicating the polarity of the answer, ranging from the gradients of \labelone\ and \labeltwo\ to middle and out-of-context labels. The meaning of the labels is explained in Table \ref{tab:label_examples} and the annotation details can be found in \S\ref{subsec:annotations_and_label_defs}.

\textbf{Language varieties and annotator demographics.}
The dataset contains data in English (en-EN), Standard German (de-DE) and a dialectal variant of German, Bavarian (de-BAR).
The Bavarian translation depicts the dialectal variant of the border area between rural Upper and Lower Bavaria. The translations were carried out by a native speaker of Standard German and Bavarian in her 20s.

\textbf{Linguistic situation and data quality.}
The English and German language as it occurs in the dataset is scripted, as it stems from movie scripts of various genres, e.g., comedy or crime. The Bavarian translation aims at being as natural sounding in the dialect as possible.

We worked with the movie subtitles as included in the opensubtitles v2018 corpus \cite{lison2016opensubtitles2018}. The sentences vary greatly in quality, as some are missing (several) words and are clearly incomplete. \inqaplus\ does keep some of these instances if the incomplete sentence makes sense as an answer by itself. 

\textbf{Data preprocessing.}
We pre-filter the opensubtitles v2018 corpus \cite{lison2016opensubtitles2018} by searching the raw German corpus for questions that are followed by a line that does not contain the direct answer particles \labelone\ and \labeltwo. We then go over the potential candidates to find question answer pairs where the answer is indirect and plausible with regard to the question. To create the English corpus, we find the parallel English sentences of the chosen German instances. The Bavarian translation is based on the English corpus as to not be influenced by the Standard German wording.

\subsection{\geniqa\ Data Statement}\label{app:geniqa_statement}

\begin{itemize}
    \item Dataset name and version: \geniqa\ (EN, DE and BAR) (version 0.1)
    \item Dataset curators: Miriam Winkler, Verena Blaschke, Barbara Plank
    \item Dataset and data statement citation: Please cite this paper.
    \item Data statement version: 1.0
\end{itemize}

\textbf{Executive summary and curation rationale.}
This dataset is a training dataset for Indirect Question Answering (IQA) resource. It contains 1.500 LLM-generated indirect question-answer pairs per language in English, Standard German and Bavarian, a German dialect.

\textbf{Date.} 
The data was generated in the time period of March to July 2025 with GPT-4o-mini \cite{openai2024gpt}.

\textbf{Modality.} Written (LLM-generated question-answer pairs).

\textbf{Genre.} Questions and indirect answers.

\textbf{Dataset statistics.}
The dataset contains of 1.500 English, Standard German and Bavarian sentences each. It consists only of a train split for each language.

\textbf{Labels.}
Each question-answer pair carries a numerical label indicating the polarity of the answer, ranging from the gradients of \labelone\ and \labeltwo\ to middle and out-of-context labels. The meaning of the labels is explained in Table \ref{tab:label_examples} and the annotation details can be found in~\S\ref{subsec:annotations_and_label_defs}.
The labels were generated by GPT-4o-mini jointly with the question--answer pairs.

\textbf{Language varieties.}
The dataset contains data in English (en-EN), Standard German (de-DE) and a dialectal variant of German, Bavarian (de-BAR). Each language was generated natively and there is no information about the regional variety of Bavarian that the model generated.

\textbf{Linguistic situation and data quality.}
The language in the dataset is scripted, as the LLM generations do not depict natural text. The generations for English and Standard German are high-quality. However, as detailed in \S\ref{subsec:geniqa_dialect_quality}, the quality of the Bavarian dialect is lacking. A large part of instances was generated in either Standard German or unnatural pseudo-dialect.

\subsection{Distribution, use, and maintenance}
\label{sec:appendix-data-distribution}

\textbf{Distribution and usage terms.}
We release the datasets in the linked GitHub repository.\footnote{\repo} \inqaplus\ can be used under the terms of opensubtitles v2018 of mentioning the corpus paper \cite{lison2016opensubtitles2018} and the source website http://www.opensubtitles.org/.

\textbf{Maintenance and updates.}
We do not plan further updates of this dataset. If you find errors, please open an issue on GitHub or contact Verena Blaschke or Barbara Plank.

\subsection{Limitations, acknowledgements, and further information}
\label{sec:appendix-data-acknowledgements}

\textbf{Limitations.}
The data format and label set corresponds to the IndirectQA \cite{mueller2024indirectqa} dataset in order to ensure compatibility. 

The Bavarian translation reflects the dialect of a single speaker.

\textbf{Disclosures and acknowledgements.}
There are no conflicts of interest.
This research is supported by European Research Council (ERC) Consolidator Grant DIALECT 101043235.

\textbf{About this data statement.}
A data statement is a characterization of a dataset that provides context to allow developers and users to better understand how experimental results might generalize, how software might be appropriately deployed, and what biases might be reflected in systems built on the software.

This data statement was written based on the template for the Data Statements Version~3 Schema. The template was prepared by Angelina McMillan-Major and Emily M.\ Bender and can be found at \href{http://techpolicylab.uw.edu/data-statements}{http://techpolicylab.uw.edu/data-statements}.

\section{\inqaplus\ Details}\label{app:inqaplus_details}
\subsection{Data Genres}\label{app:inqa_genres}

\begin{figure*}[]
    \centering
    \begin{subfigure}[t]{0.3\textwidth}
         \centering
         \includegraphics[width=\textwidth]{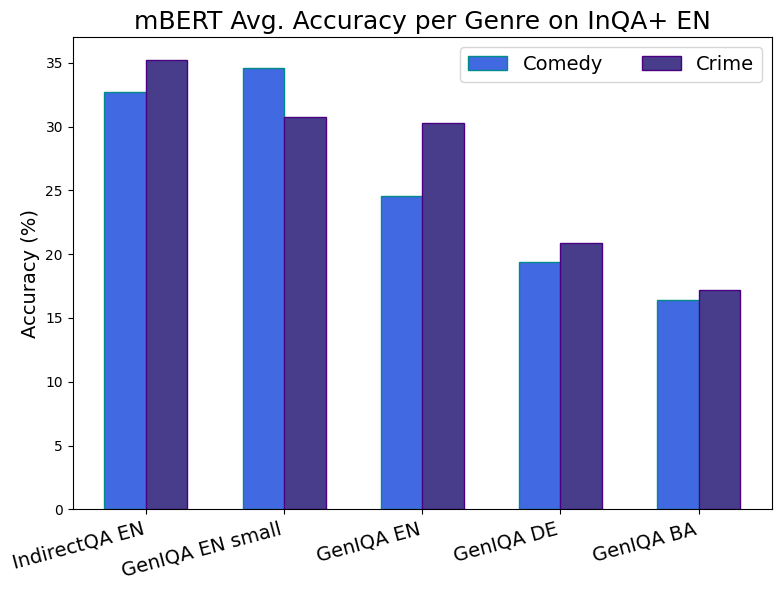}
         \caption{English.}
         \label{fig:genres-mbert-en}
     \end{subfigure}
     \hfill
     \begin{subfigure}[t]{0.3\textwidth}
         \centering
         \includegraphics[width=\textwidth]{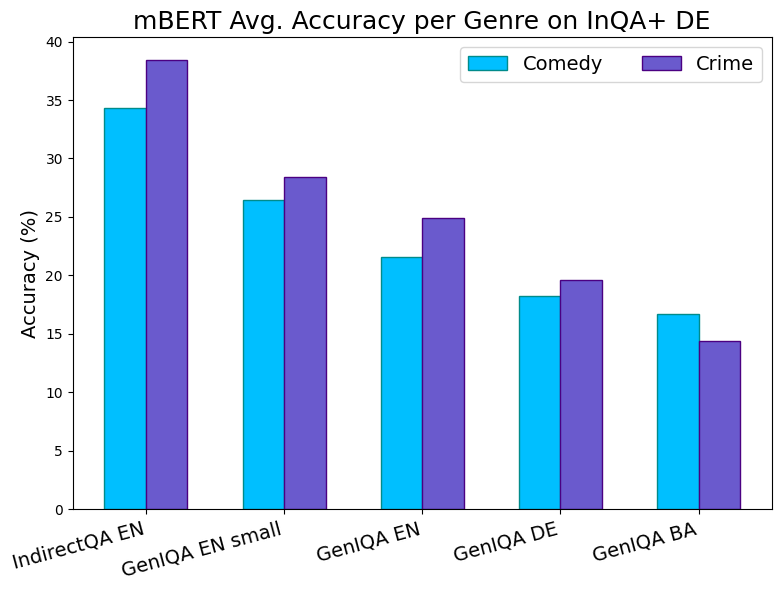}
         \caption{German.}
         \label{fig:genres-mbert-de}
     \end{subfigure}
     \hfill
     \begin{subfigure}[t]{0.3\textwidth}
         \centering
         \includegraphics[width=\textwidth]{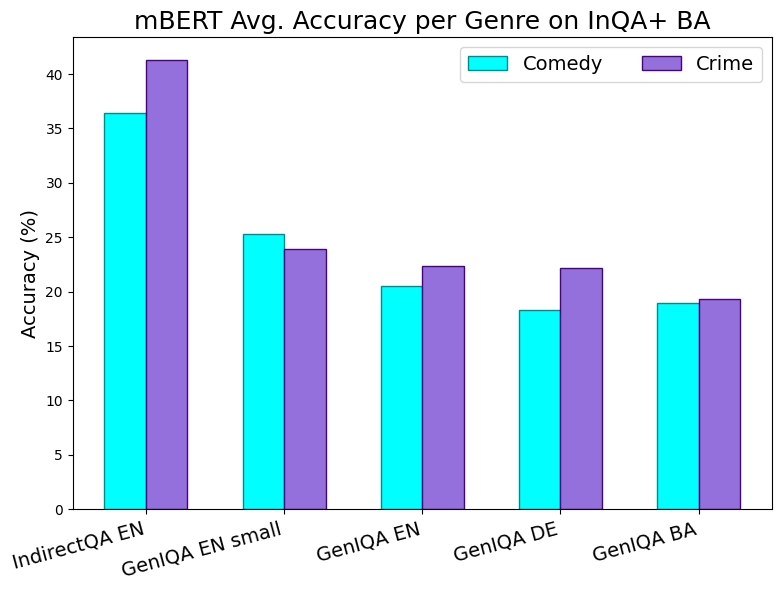}
         \caption{Bavarian.}
         \label{fig:genres-mbert-ba}
     \end{subfigure}
    \caption{Average accuracy scores per genre over three seeds of mBERT models, evaluation on \inqaplus.}
    \label{fig:genres-mbert}
\end{figure*}

\begin{figure*}[]
    \centering
    \begin{subfigure}[t]{0.3\textwidth}
         \centering
         \includegraphics[width=\textwidth]{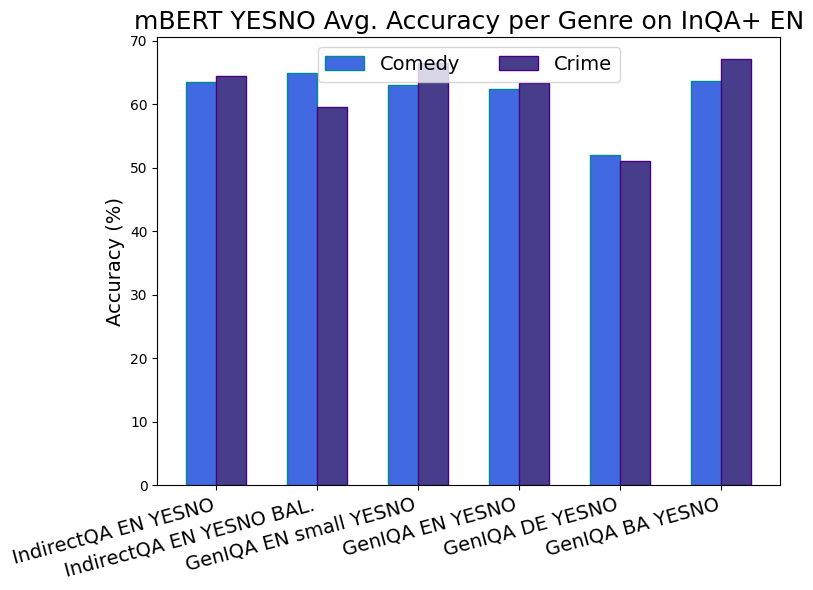}
         \caption{English.}
         \label{fig:genres-mbert-en}
     \end{subfigure}
     \hfill
     \begin{subfigure}[t]{0.3\textwidth}
         \centering
         \includegraphics[width=\textwidth]{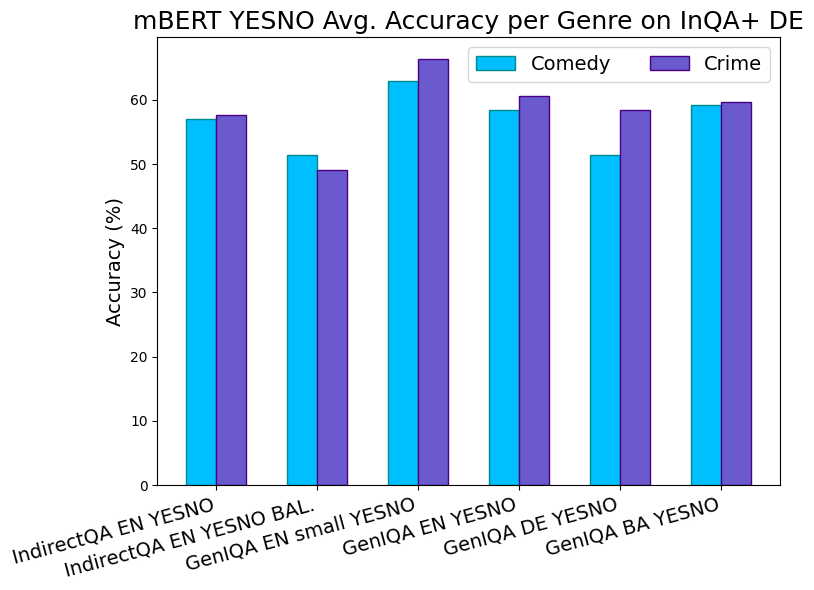}
         \caption{German.}
         \label{fig:genres-mbert-de}
     \end{subfigure}
     \hfill
     \begin{subfigure}[t]{0.3\textwidth}
         \centering
         \includegraphics[width=\textwidth]{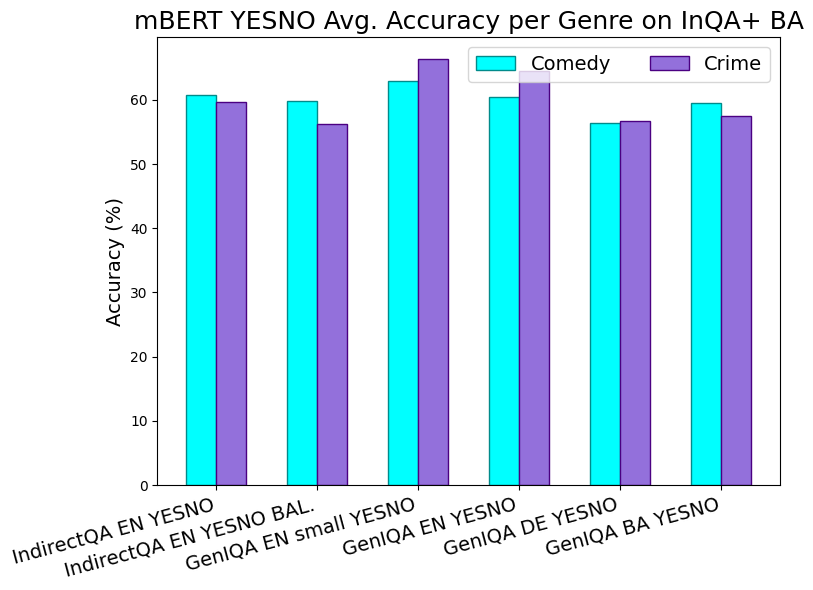}
         \caption{Bavarian.}
         \label{fig:genres-mbert-ba}
     \end{subfigure}
    \caption{Average accuracy scores per genre over three seeds of mBERT models, evaluation on \inqaplus\ \yesno.}
    \label{fig:genres-yesno-mbert}
\end{figure*}

\citet{mueller2024indirectqa} divide their dataset IndirectQA into two separate genres: comedy and crime to research the effect of domain
They found that comedy sentences, that contain complex pragmatic structures like sarcasm %
more frequently than crime %
yield lower accuracy scores.

We do not see distinctive differences in performance when dividing \inqaplus\ into comedy and crime test sets and evaluating our trained models on the separate genres. However, the accuracy of the crime test set is generally slightly higher (Figure \ref{fig:genres-mbert}), fitting into the observation by \citet{mueller2024indirectqa}. The same trend applies to \inqaplus\ \yesno\ (Figure \ref{fig:genres-yesno-mbert}), even if the disparities are even lower.

\subsection{Annotation Details}\label{app:inqa_annotation_details}
In \S\ref{subsec:annotations_and_label_defs}, we explain the annotation %
of \inqaplus. The label definitions %
are the main annotation guideline%
with an additional rule for negative questions %
that are a challenge for human and machine annotations.
With the additional rule, negative questions can be disambiguated and annotated uniformly across larger datasets. However, while this method untangles the polarity for humans to understand, the uncertainty of the answers' meanings %
still is a key factor to the difficulty of the IQA task. We interpret the labels as following:

\begin{itemize}
    \item \textbf{\labelone}: \textbf{A clear yes or all gradients of yes (including weaker forms, e.g., maybe yes).} Every instance that answers the question positively and confirms the statement. Clear \labelone\ answers include instances like \textit{I sure do} or \textit{Of course}, while the gradients are more varied, for example \textit{I guess so} or \textit{Oh, well, maybe}.
    
    \item \textbf{\labeltwo}: \textbf{A clear no or all gradients of no.} Every instance that answers the question negatively and contradicts the statement. Clear \labeltwo\ answers include sentences like \textit{Oh, I can't} or \textit{Course not}, while the gradients consist of instances like \textit{Not yet, but I will} or \textit{Probably not}.

    \item \textbf{\labelthree}: \textbf{A yes that only holds if certain conditions are true.}  Every instance that answers the question positively, but there is a direct or indirect condition. %
    For example, \textit{I will be if you want me to be} is a direct condition.
    Indirect conditions are inferred from the answer, like in %
    \textit{You going by Audrey? -- I can.} 
    -- meaning yes, the responder can, if the questioner wants.

    \item \textbf{\labelfour}: \textbf{A neutral answer that lies in the middle of yes and no.}  Every instance that does answer the question, but the polarity is not clearly positive or negative. In the majority of cases %
    the responder is thinking (e.g., \textit{Let me think about that for a second}) or doesn't know the answer (\textit{I don't know}). 

    \item \textbf{\labelfive}: \textbf{The sentence does not match the questions as an answer.} Every instance that does not fit the question, because there is no clear connection without knowing further context, for example: \textit{Another trick question, right? --Open the window}. 
    We also label short phrases like \textit{What?} or \textit{Hm?} as \labelfive, as the answers do not carry informative meaning.

    \item \textbf{\labelsix}: \textbf{Without further context, the answer cannot be clearly categorized.} Every instance where the polarity depends on the context around the question-answer pair, e.g., \textit{Does he want us to go? -- He was very quiet on the phone}. Without knowing the person referenced, it is unclear whether \textit{he} wants the attendance or not. This label also includes sarcastic answers like \textit{You would've taken the subway? -- What do you think?}. For more information on the difference between \labelfour, \labelfive\ and \labelsix, please refer to \S\ref{subsec:annotations_and_label_defs}.
\end{itemize}

\paragraph{Complex phenomena: negative questions}
In \inqaplus, 31 IQA pairs (7\,\% of the total instances) contain negative questions, which raises the need to standardize the annotation of such cases. The interpretation is context- and speaker dependent \cite{duskova1981negativequestions}.

As explained in \S\ref{subsec:annotations_and_label_defs}, we resolve negative questions with polarity-based interpretations \cite{holmberg2012syntax, holmberg2013syntax}, as it is clearer to understand and annotate for various annotators. When calculating the IAA on a very small sample of double-annotated instances, we get an agreement of 0.35 Cohen's Kappa score before setting the additional guidelines, which is interpreted as no or only slight agreement. After streamlining the interpretation of negative questions as polarity-based, we raise the agreement to 0.54, falling in line with the agreement on the whole dataset.

The question \textit{Are you not Moby?} has three different possible answers: two \labelone\ answers -- \textit{(Yes) I am Moby} (polarity-based) or \textit{(Yes) I am not Moby} (truth-based) -- and one \labeltwo\ answer -- \textit{(No) I am not Moby}. For German speakers, the \labelone\ answers can be disambiguated with the the word \textit{Doch}, an adverb that has no English equivalent, as a help for decision making. In our example case, it carries the meaning of \textit{Yes, I am (indeed) Moby} and cannot be confused with \textit{Yes, I am not Moby}.

Further, in the case of IQA, the answer additionally might not always clearly reveal the polarity like in the \textit{Moby} example. Looking at the question-answer pair \textit{Can’t you hear the bell ringing? - That’s Lilly’s job}, the annotators additionally have to decode the indirectness in the answer before annotating. The answer implies that the bell was heard, but that the responding person does not feel responsible to take any action. With this inferred knowledge, the more direct answer can be decoded as \textit{(Yes) I can hear the bell ringing, but that is Lilly’s job}, leading to a \labelone\ annotation.

\section{LLM Generation Tests}\label{app:llm_generation_tests}
\begin{table*}[h!]
    \centering
    \resizebox{\textwidth}{!}{%
    \begin{tabular}{l|ccccc}
    \toprule
        Model Name & Instruction-tuned? & Dialect? & Correct? & Logical? & Indirect?  \\ \midrule
        EuroLLM-9B-Instruct & \checksmall & \checksmall & \checksmall & \checksmall &  \checksmall \\
        EuroLLM-1.7B-Instruct & \checksmall & \crosssmall & \crosssmall & \crosssmall & \crosssmall \\
        OLMo-2-1124-7B-Instruct & \checksmall & \checkyellow & \checksmall & \checkyellow & \checksmall \\
        gemma-2-2b-it & \checksmall & \checkyellow & \checksmall & \checksmall & \checkyellow \\
        Llama-3.2-1B-Instruct & \checksmall & \crosssmall & \crosssmall & \checkyellow & \checkyellow  \\
        gpt-4o-mini & \checksmall & \checksmall & \checksmall & \checksmall & \checksmall  \\
        \midrule \midrule
        LLäMmlein\_1B & \crosssmall & / & / & / & /  \\
        Betzerl\_1B\_wiki\_preview & \crosssmall & / & / & / & / \\
    \bottomrule
    \end{tabular}%
    }
    \caption{Assessment matrix of relevant features for IQA and Bavarian for various LLMs that are true (\checksmall), partially true with some restrictions (\checkyellow) or false (\crosssmall).}
    \label{tab:llmtesting}
\end{table*}

The selection of the right model is crucial for the generation quality which is why we took great care in testing the models. 
As explained in \S\ref{subsec:geniqa}, we focus our testing on the generation of Bavarian, a German dialect.

\subsection{LLM dialect generation testing}
The generation of authentic, natural sounding Bavarian is not trivial.
We first test the models' dialect generation capabilities with the prompt \textit{Write a sentence in Bavarian dialect} before presenting the models with the complex IQA task (\textbf{Dialect} criterion in Table \ref{tab:llmtesting}). 

The only free model capable of producing natural sounding, comprehensible Bavarian is EuroLLM-9B \cite{martins2025eurollm}. 
It returns sentences like \textit{I bin do in Land, wo’s Bier so g’scheit is} and also provides the unprompted translation \textit{I am in the land where beer is so good}. 
However, its run time of a minimum of 5+ hours on the computational resources we had available for this study was too long to be feasible for producing a whole dataset.
Other than EuroLLM-9B, GPT-4o-mini \cite{openai2024gpt} as a paid API also generated natural sounding dialect. The other LLMs generated mostly ``gibberish'' sentences.

We continue with zero- and few-shot prompts tailored to the IQA task and our label set with IQA examples (for few-shot). From the generation results, we analysed if the sentences were coherent and complete (\textbf{Correct} criterion), if the answer made sense with regards to the question (\textbf{Logical} criterion) and if the answers were indirect (\textbf{Indirect} criterion) in Table \ref{tab:llmtesting}. In general, the best results were acquired within the few-shot scenario. The results of the zero-shot prompting were often arbitrary and did not follow the given instructions.

The highest quality results were generated by GPT-4o-mini. It follows the prompt in detail and generates coherent dialect. Even though OpenAI is not open about their training data sources, we can see from the generation performance that GPT-4o-mini has seen Bavarian text in the supposed assortment of openly available corpora, scraped internet text and reinforced post-learning through human feedback \cite{openai2024gpt}.

Even though we only see unfinished generations with EuroLLM-9B due to interruptions, the experiments with a few-shot IQA prompt show the good dialect quality: \textit{Hast denn da Frida schon am Bus-Halt gwunn} (translation with assumptive completion in []: \textit{Have you [waved] to Frida at the bus stop yet?}). The article \textit{da} in front of the proper name and the omitted \textit{e} in between \textit{g} and \textit{w} in \textit{gwunn} -- in contrast to Standard German prefix \textit{ge-} -- is typical for the Bavarian dialect.
While EuroLLM-9B was very promising, the smaller version with faster run times, EuroLLM-1.7B \cite{martins2025eurollm}, did neither produce dialect nor follow the prompt instructions. 

OLMo-2 \cite{olmo20242olmo2furious} had a similarly low performance: the generated answers did to the most parts contain no dialect and instead of following the prompt, it either translates or just returns parts of the requested question-answer-label format, e.g., the label.
Gemma-2 \cite{gemmateam2024gemma2} produced pseudo-Bavarian. The sentence structure and question-answer logic is correct, but the dialect does not make any sense. Llama’s \cite{meta2024llama} production is worse than Gemma’s as it only produced gibberish with the same sentence structure of \textit{Do you have [...] - I have [...]}. 

Unfortunately, we could not test generating with the German LLM LLäMmlein\_1B \cite{pfister2024llammlein} and the Bavarian adapter Betzerl \cite{caidas2025betzerl} as they are not instruction-tuned and thus not suitable for our purpose.

\subsection{Prompt wording testing}
For the generation of the full \geniqa\ datasets, we fine-tuned the prompt to better capture the beneficial features provided by OpenAI’s best practices \cite{openai2024gpt}. The following prompt is the final one we used for the generation process:\\

\resizebox{\linewidth}{!}{
\begin{tabular}{c}
    \textit{\makecell[l]{Write 50 polar questions and sentences that answer the\\ questions. Both should be in English and formatted as\\ the given format. The answer should be indirect, \\meaning that it does not contain direct answer particles\\ like Yes, No or similar indicators. The answer should be\\ labelled with one of the given labels from the label set \\and therefore be suitable for the explanation of the\\ label. Focus on the labels 1 to 4, but also include\\ examples for the labels 5 and 6. Do not number \\the examples, just write them in the given format.\\ \\Desired format: question answer label\\ \\}}
\end{tabular}
}\\

\resizebox{\linewidth}{!}{
\begin{tabular}{c}
    \textit{\makecell[l]{Label set:\\1 = Yes: A clear yes or all gradients of yes (meaning \\weaker forms of yes like a maybe yes).\\2 = No: A clear no or all gradients of no.\\3 = Conditional Yes: A yes that only holds if certain \\conditions are true.\\4 = Neither yes nor no: A neutral answer that lies in the \\middle of yes and no.\\5 = Other : The sentence does not match the question\\ as an answer.\\6 = Lacking context: Without further context, the answer\\ cannot be categorized as yes or no.\\ \\Example for each label:\\This is Long Tieng, right? Right. 1\\You hungry? Just black coffee for me. 2\\Will you be home tonight? I will be if you want me to be. 3\\Is anyone still in there? I don't know. 4\\Another trick question, right? Open the window. 5\\- Hey, you guys want sangria? - It’s Guys’ Night! 6}}
\end{tabular}
}\\

The prompt contains more fine-grained instructions than our testing prompts and explains the meaning of the label sets clearer. As we saw the best results with few-shot prompting, we added more examples (one for each label) from \inqaplus.
Additionally, we tell the model to focus on labels \labelone, \labeltwo, \labelthree\ and \labelfour, all of which bear the most interesting information for the task. Without this addition, the model had the tendency to label most generations as \labelfive. 

\paragraph{Prompt language testing}
Since we experiment with the data creation for EN, DE and BAR, we also try out in-language prompts, meaning a German and a Bavarian prompt for the creation of Bavarian data. With the idea that the in-language prompts ``prime'' the model for a closer language to the one we want to generate, we translate our IQA generation prompt above accordingly. 
For the German split of \geniqa, the prompt language did not make a big difference: the labelling accuracy of 100 sentences generated with the German prompt is 44\,\% -- comparable to the 45\,\% reached with the English prompt.
The Bavarian prompt did not work well and yielded poor generations.

\subsection{Generation temperature testing}
After fine-tuning the prompt, we experiment with the generation temperature.
It is an important parameter for the quality of generations as a high temperature makes the generation more random and diverse and a lower temperature makes them more deterministic with the chance of it being more repetitive \cite{openai2024gpt}.
We want creative and varied sentences for the \geniqa\ datasets but keep the labelling accuracy of the generated annotations as high as possible.
Thus, we compare 100 generations with the temperatures 0.2 and 0.8 respectively and manually calculate the labelling accuracy by hand-annotating the sentences. For the low temperature, we find an agreement of 48\,\%; with the high temperature, it lies at 59\,\%. Higher temperature thus gives more create freedom and higher labelling accuracy to the model, which is why we choose this value.

\section{Generated Dialect Quality}\label{app:generated_dialect_quality}
We provide more details on the dialect quality of the LLM-generated Bavarian data (\geniqa\ BAR) which we discussed in \S\ref{subsec:geniqa_dialect_quality}. We first present a more in-depth look into our manual quality assessment and provide the details of our native speaker survey afterwards.

\begin{figure*}[]
    \centering
    \begin{subfigure}[t]{0.32\textwidth}
         \centering
         \includegraphics[width=\textwidth]{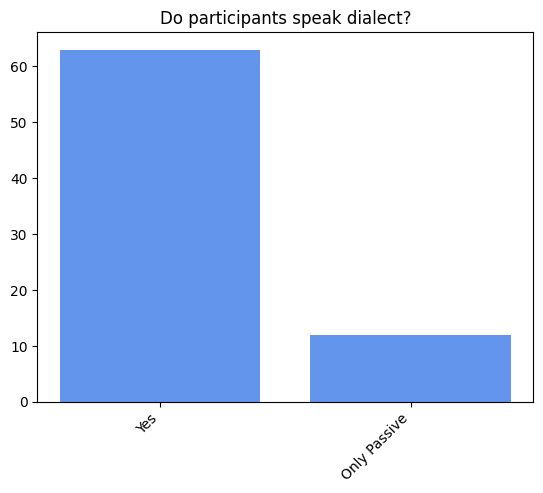}    
         \caption{Information about \textit{if} participants speak dialect.}
         \label{fig:survey-dialect-skills-speak}
     \end{subfigure}
     \begin{subfigure}[t]{0.32\textwidth}
         \centering
         \includegraphics[width=\textwidth]{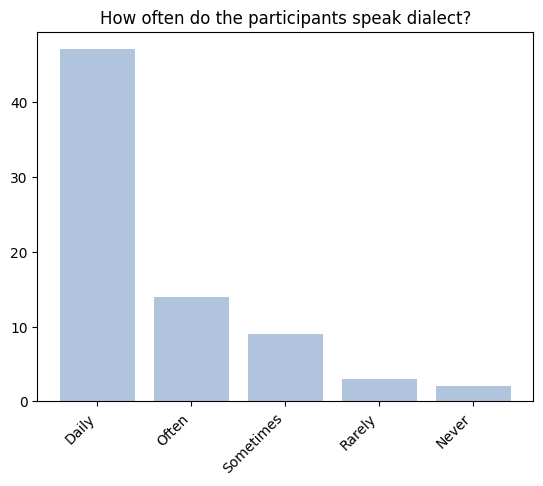} 
         \caption{Information about \textit{how often} participants speak dialect.}
         \label{fig:survey-dialect-skills-often}
     \end{subfigure}
     \hfill
     \begin{subfigure}[t]{0.32\textwidth}
         \centering
         \includegraphics[width=\textwidth]{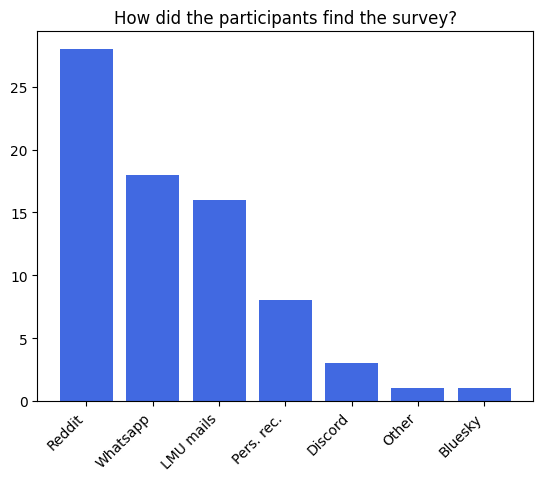}
         \caption{Distribution channels where the participants found the survey.}
         \label{fig:survey-origin-source}
     \end{subfigure}
    \caption{Personal disclosures of the participants in the dialect quality survey.}
\end{figure*}

\subsection{Manual dialect quality assessment of \geniqa\ BAR}
We want to analyse the quality of \geniqa\ BAR as generated by GPT-4o-mini \cite{openai2024gpt} and differences between artificial and natural data for low-resource languages in more detail. 
The model does not use Bavarian grammar, dialect vocabulary (words specific in the dialect that does not exist/is not used in the standard language) or dialect spellings that match the pronunciation consistently.

We noticed two prominent positive features.  %
Firstly, in dialect speech, clitics are characteristic and common \cite{weiss1998bavariangrammar, blaschke2024maibaamguidelines}, e.g., merging the word \textit{es} (it) into the preceding words. The model applies this for example in the sentences \textit{I find die Berge schön, aber am Meer \textbf{is’s} auch nett} (I think the mountains are beautiful, but it’s also nice by the sea). %
Secondly, the best sounding artificial dialect sentences contain words with elisions (e.g., missing vowels, especially prevalent in pre- and suffixes), like \textit{g’wesn} instead of \textit{gewesen} in \textit{I bin da schon oft \textbf{g’wesn}} (I have been there a lot).

However, the most generations contain erroneous or pseudo dialect (incorrectly imitated dialect characteristics).
We classify three prominent error classes: wrong usage of verbs, incorrect omissions and determiners of false genus. Wrong usage of verbs refers to verbs that do not fit into the grammatical context. 
For example, the answer \textit{I \textbf{hätt} gern, aber es gibt viel zu tun} (I would like to, but there is a lot do do) uses \textit{hätt} (subjunctive II form of the verb \textit{haben} (to have)) that does not suit its question \textit{Kommst du mit uns zum Grillen?} (Are you coming with us to the barbecue?). Subjunctive II of \textit{werden} (would) would be correct instead. %

Then, GPT-4o-mini replicates clitics and elisions, but produces incorrect letter omissions. In the sentence \textit{I hab a paar Blüm’ und Kräuter, aber ned viel Platz} (I have a few flowers and herbs, but not much space), the word \textbf{Blüm’} (flower) does not exist in this form in the dialect of the annotator and is wrongly cut short. 

False genus generations are also common in \geniqa\ BAR.  For example, \textit{I ess sie, wenn’s an Fest gibt} (I eat them when there’s a party) uses \textit{an}, the accusative masculine determiner. As \textit{Fest} is grammatically neuter, the correct dialect determiner is \textit{a}. %
This behaviour is not restricted to dialect sentences and also sometimes occurs with Standard German.

Additionally, we observed in the sample of 100 instances we analysed in \S\ref{subsec:geniqa_dialect_quality} %
that the kind of vocabulary GPT-4o-mini generates in Bavarian differs from Standard German.
Many \geniqa\ BAR sentences contain stereotypical Bavarian words that each have zero occurrences in \geniqa\ DE but multiple in the whole Bavarian data, for example \textit{Bier} (beer), \textit{Wirtshaus} (traditional Bavarian restaurant) and traditional Bavarian food like \textit{Leberkäse} (meat loaf) and \textit{Brezen} (pretzels) are a selection of the words we found. Since OpenAI does not disclose their training data in detail, we assume that the dialect data GPT-4o-mini encountered strongly covered Bavarian traditions to build the connection of this vocabulary and Bavarian.

\subsection{Native speaker survey}

Bavarian dialect is a broad and varied spectrum with many regional varieties. We conduct an anonymous survey and ask multiple native Bavarian speakers from different regions in Bavaria to rate the quality of the AI-generated dialect, as the perception of the dialect may vary depending on the participants' own spoken dialect variety. 
For this, we ask the participants to disclose the administrative region (Upper Bavaria, Lower Bavaria, Upper Palatinate, Upper Franconia, Middle Franconia, Lower Franconia or Swabia) they live in as the only personal information.
The online survey was conducted with SoSci Survey \cite{leiner2025sosci} and the data was collected between 29.05.2025 and 11.06.2025. 
In total, 76 active dialect speakers of varying degrees (Figure \ref{fig:survey-dialect-skills-speak}) answered the survey, with the majority of them stemming from Upper and Lower Bavarian (Figure \ref{fig:survey-origin-region}) and speaking dialect on a daily basis (Figure \ref{fig:survey-dialect-skills-often}). Out of the seven distribution channels, the most participants found the survey on Reddit %
(Figure \ref{fig:survey-origin-source}). 

The design of the survey is independent from the IQA task to assess only the linguistic quality of the generated dialect and to allow participants without linguistic knowledge to partake. The participants' task is to judge ten question-answer pairs on if the answer contains dialect and if it is authentic (meaning that they would use such an answer in their daily life).
We select the examples in a way that each represents a different generation phenomena and degree of quality (low to high):

\begin{figure}
  \centering
    \includegraphics[width=\linewidth]{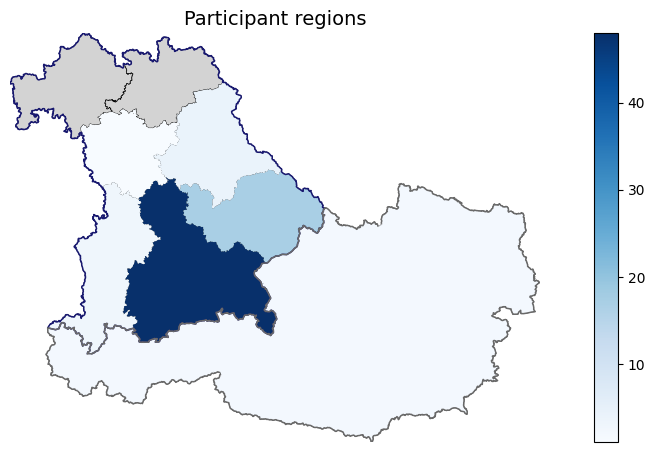}
  \caption{Participant origin regions of the dialect quality survey: Bavaria (per administrative region) and Austria.}
  \label{fig:survey-origin-region}
\end{figure}

\begin{figure*}
    \centering
    \begin{subfigure}[t]{0.49\textwidth}
         \centering
         \includegraphics[width=\textwidth]{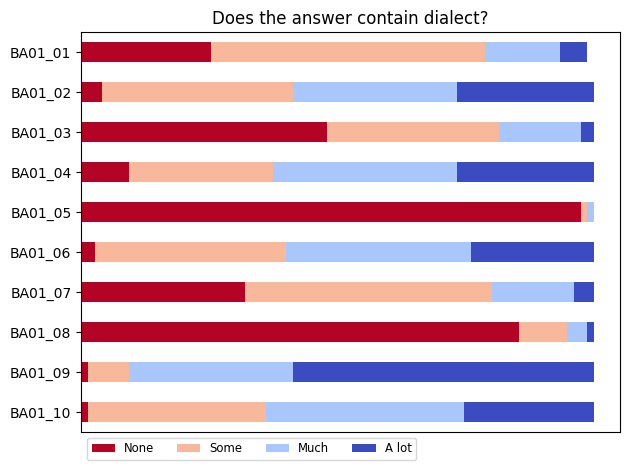}
         \caption{Dialect presence.}
         \label{fig:survey_dialect_presence}
     \end{subfigure}
     \hfill
     \begin{subfigure}[t]{0.49\textwidth}
         \centering
         \includegraphics[width=\textwidth]{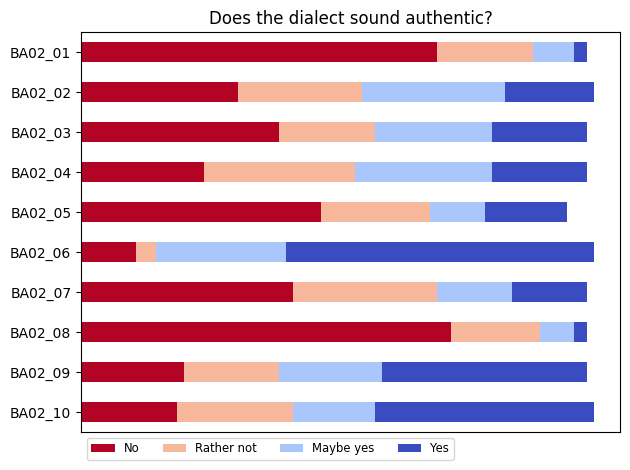}      
         \caption{Dialect authenticity.}
         \label{fig:survey_dialect_auth}
     \end{subfigure}
    \caption{Results of the dialect questions.}
    \label{fig:survey_dial_questions}
\end{figure*}

\begin{figure*}
    \centering
    \begin{subfigure}[t]{0.49\textwidth}
         \centering
         \includegraphics[width=\textwidth]{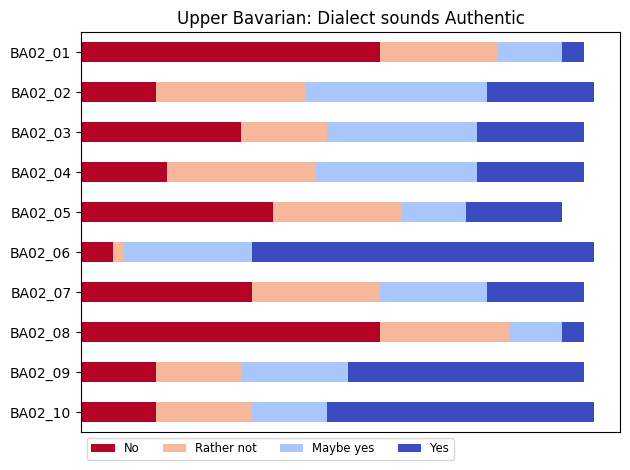}
         \caption{Authenticity rating in Upper Bavaria (author's region).}
         \label{fig:survey_dialect_auth_upper}
     \end{subfigure}
     \hfill
     \begin{subfigure}[t]{0.49\textwidth}
         \centering
         \includegraphics[width=\textwidth]{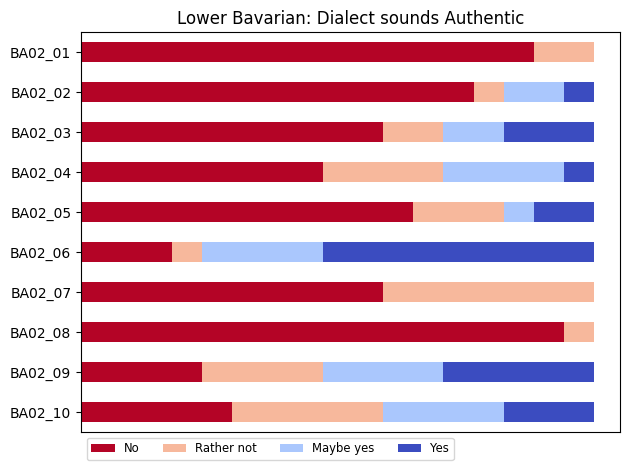}      
         \caption{Authenticity rating in Lower Bavaria.}
         \label{fig:survey_dialect_auth_lower}
     \end{subfigure}
    \caption{Results of the dialect questions per region. }
    \label{fig:survey_dialect_auth_lower_higher}
\end{figure*}

\begin{enumerate}
    \item Kann ich bei dir lernen? - So lange du ned sabbelst, klar. \\ \textit{\textcolor{gray}{Can I learn at your place? - As long as you don’t blabber, sure.}} \\ The word \textit{sabbeln}\footnote{\href{https://www.dwds.de/wb/sabbeln}{https://www.dwds.de/wb/sabbeln}} (blabber) stems from Northern German dialects. \\ \textit{Quality: Low}
    \item Hast du den Film schon gesehen? - I hob nur die Werbung gseh. \\ \textit{\textcolor{gray}{Have you seen the film yet? - I’ve only seen the ad.}} \\ Contains pseudo-dialect (\textit{gseh}, eng. seen). \\ \textit{Quality: Low}
    \item  Ist der Chef im Büro? - Möglicherweise is er unterwegs.  \\ \textit{\textcolor{gray}{Is the boss in the office? - He may be on the move.}} \\ Only contains one dialect word (\textit{is}, eng. is). \\ \textit{Quality: Medium}
    \item  Wirst du morgen aufstehen? - I schau ma amoi, wies mir geht.  \\ \textit{\textcolor{gray}{Will you get up tomorrow? - I’ll see how I’m doing.}} \\ Wrong Bavarian grammar (\textit{ma} (Bavarian version of \textit{we}) exposes the sentence as fake dialect as it is incorrect here). \\ \textit{Quality: Low}
    \item  Ist die Pizza fertig? - Es riecht schon ganz lecker! \\ \textit{\textcolor{gray}{Is the pizza ready? - It already smells delicious!}} \\ Standard German. \\ \textit{Quality: Low}
    \item  Habt ihr einen Tisch reserviert? - I hab’s vergessen.  \\ \textit{\textcolor{gray}{Did you book a table? - I forgot.}} \\ Only contains one dialectal word, but sounds authentic nonetheless. \\ \textit{Quality: High}
    \item  Gehst du oft ins Kino? - Ich schau lieber Filme doheim.  \\ \textit{\textcolor{gray}{Do you often go to the cinema? - I prefer to watch films at home.}} \\ Wrong dialect spelling (\textit{doheim}, std. ger. daheim, eng. at home). \\ \textit{Quality: Low}
    \item  Ist das Bier kalt? - Es steht’s im Kühlschrank.  \\ \textit{\textcolor{gray}{Is the beer cold? - It’s in the fridge.}} \\ No expression of the intended meaning. The answer is Bavarian, but only with the interpretation of \textit{You (plural) are in the fridge}. For the question, it does not hold the correct meaning. \\ \textit{Quality: Low}
    \item  Hast du schonmal Ente gekocht? - I kimm mid de Entn klar. Entspann di.  \\ \textit{\textcolor{gray}{Have you ever cooked duck? - I can handle the ducks. Relax.}} \\ Manually translated answer by the annotator \\ \textit{Quality: High}
    \item  Isst du etwas Gesundes? - I hätt grad a Lust auf a Stück Pizza. \\ \textit{\textcolor{gray}{Are you eating something healthy? - I crave a slice of pizza right now.}} \\ Uncommon Bavarian grammar (using an article with \textit{Lust} (craving) is not wrong per se, but unusual). \\ \textit{Quality: Medium}
\end{enumerate}

\begin{table*}[]
    \centering
    \begin{tabular}{l|rrrrrr}
    \toprule
        \makecell[l]{Model\\ \qquad Train Data} & \makecell[c]{LR} & \makecell[c]{Batch\\ Size} & \makecell[c]{Warm-up \\Ratio} & \makecell[c]{Weight \\Decay} & \makecell[c]{Dropout\\Rate} & \makecell[c]{Label\\Smoothing} \\ \midrule
        mBERT & & & & & & \\
        \qquad IndirectQA EN & 1e-5 & 16 & 0.1 & 0.1 & 0.1 & 0.0 \\
         \qquad \geniqa\ EN & 1e-5 & 16 & 0.5 & 0.01 & 0.5 & 0.1 \\
         \qquad \geniqa\ EN \textsc{small} & 1e-5 & 8 & 0.3 & 0.01 & 0.5 & 0.1 \\
         \qquad  \geniqa\ DE & 1e-5 & 8 & 0.1 & 0.001 & 0.5 & 0.3 \\
         \qquad  \geniqa\ BA & 1e-5 & 4 & 0.1 & 0.01 & 0.5 & 0.1 \\ 
         \qquad IndirectQA EN \yesno & 1e-5 & 16 & 0.1 & 0.1 & 0.1 & 0.0 \\
        \qquad \geniqa\ EN \yesno & 1e-5 & 4 & 0.1 & 0.001 & 0.3 & 0.1 \\
        \qquad \geniqa\ DE \yesno & 1e-4	& 16 &  0.3  & 0.1  &  0.1 & 0.0  \\
         \qquad \geniqa\ BAR \yesno & 1e-5 & 16 & 0.3 & 0.1 & 0.1 & 0.0 \\[2ex]
        mDeBERTa & & & & & & \\
        \qquad IndirectQA EN & 1e-4 & 32 & 0.1  & 0.1 & 0.1 & 0.0 \\
         \qquad  \geniqa\ EN & 1e-4 & 32  & 0.1  & 0.01 & 0.1  & 0.3  \\
         \qquad \geniqa\ DE & 1e-5 & 32  & 0.1  &  0.001 & 0.1  & 0.1  \\
         \qquad  \geniqa\ BAR & 1e-4 & 4 & 0.5 & 0.1 & 0.1 & 0.0 \\ [2ex]
        XLM-R & & & & & & \\
        \qquad IndirectQA EN & 1e-5 & 32 & 0.5 & 0.1 & 0.5 & 0.3 \\ 
         \qquad  \geniqa\ EN & 1e-5 & 32 & 0.3 & 0.1 & 0.1 & 0.0 \\
         \qquad  \geniqa\ DE & 1e-5 & 8 & 0.1 & 0.01 & 0.1 & 0.1 \\
         \qquad  \geniqa\ BA & 1e-5 & 16 & 0.3 & 0.1 & 0.1 & 0.0  \\
    \bottomrule
    \end{tabular}%
    \caption{Grid-searched parameters per model and training dataset after fine-tuning three seeds each for extensive experiments.}
    \label{tab:hyperparams}
\end{table*}

\begin{table*}[]
    \centering
    \resizebox{\textwidth}{!}{
    \begin{tabular}{l|c|rrrrrr}
    \toprule
        \makecell[l]{Model\\ \qquad Train Data} & Trials & \makecell[c]{LR} & \makecell[c]{Batch\\ Size} & \makecell[c]{Warm-up \\Ratio} & \makecell[c]{Weight \\Decay} & \makecell[c]{Dropout\\Rate} & \makecell[c]{Label\\Smoothing} \\ \midrule
        mBERT \\
        \qquad IndirectQA EN \remapped & 30 & 4.4274e-4 & 32 & 0.1587 & 0.0015 & 0.4748 & 0.1929 \\ 
        \qquad IndirectQA EN + Friends-QIA \remapped &  30 & 1.6414e-6 & 4 & 0.1747 & 0.0524 & 0.3930 & 0.0753 \\ [1ex]
        
        \qquad Friends-QIA & 30 & 2.1156e-6 & 8 & 0.4961 & 0.0065 & 0.3877 & 0.0931 \\
        \qquad IndirectQA EN + Friends-QIA & 30 & 4.2071e-6 & 32 & 0.1554 & 0.0003 & 0.1300 & 0.0706 \\
    \bottomrule
    \end{tabular}%
    }
    \caption{Random-searched parameters over 30 trials for mBERT per training dataset for exploratory experiments. Rounded to four digits.}
    \label{tab:hyperparams_random}
\end{table*}

As the ratings in Figure \ref{fig:survey_dialect_presence} reveal, the participants recognized the Standard German and natively translated answer from \inqaplus\ BAR. For answers 1, 3 and 7, the participants’ opinions were rather negative. %
Answer 8's rating is very negative. Only 5 out of 10 answers (2, 4, 6, 9 and 10) were rated to contain dialect. As we already touched upon in Section \ref{subsec:geniqa_dialect_quality}, the ratings from dialect speakers stemming from Upper Bavaria -- the author's region -- are higher than from Lower Bavaria (cf. Figure \ref{fig:survey_dialect_auth_lower_higher}). We distinguish only Upper and Lower Bavaria due to not having enough participants from other regions to achieve representative and comparable results.

From this selection, we see that the answers received uniformly mixed ratings for the authenticity of the dialect, except for answer 6, as visualised in Figure \ref{fig:survey_dialect_auth}. As we expected, this answer is mainly rated as authentic, meaning that the participants would use this sentence in their daily lives.

\section{Hyperparameter Fine-tuning}\label{app:hyperparameter_finetuning}
\begin{table*}[h!]
    \centering
    \resizebox{\textwidth}{!}{
    \begin{tabular}{l|ccc|c}
    \toprule
        \makecell[l]{Dataset \\ \qquad Training Setup} & \makecell[c]{Train \\Accuracy} & \makecell[c]{Dev \\Accuracy} & Gap & \makecell[c]{Majority Class \\Baseline}  \\ \midrule
        \textit{mBERT trained with} & & & \\
        \qquad IndirectQA EN & 53.71 \stdev{5.71} & 43.90 \stdev{4.53} & 9.81 & 35.61 \\
        \qquad \geniqa\ EN  & 60.78 \stdev{2.85} & 58.56 \stdev{4.60} & 2.22 & 31.80 \\
        \qquad \geniqa\ DE  & 56.82 \stdev{8.09} & 51.22 \stdev{2.52} & 5.60 & 30.93 \\
        \qquad \geniqa\ BAR  & 61.31 \stdev{1.00} & 52.78 \stdev{3.24} & 8.53 & 31.20 \\ [1ex]
        
        \qquad IndirectQA EN \textsc{remapped}  & 53.50 \stdev{4.40} & 51.76 \stdev{2.86} & 1.73 & 37.56 \\ 
        \qquad IndirectQA EN + Friends-QIA \textsc{remapped}  & 56.65 \stdev{1.40} & 56.03 \stdev{1.49} & 0.62 & 47.54 \\ [1ex]

        \qquad IndirectQA EN \yesno & 82.27 \stdev{7.84} & 74.44 \stdev{2.55} & 7.83 & 73.24 \\
        \qquad \geniqa\ EN \yesno\ & 88.18 \stdev{4.84} & 84.89 \stdev{0.84} & 3.29 & 67.73 \\
        \qquad \geniqa\ DE \yesno\ & 82.79 \stdev{3.70} & 78.96 \stdev{1.50} & 3.82 & 59.76 \\
        \qquad \geniqa\ BAR \yesno\ & 87.43 \stdev{2.68} & 82.72 \stdev{3.99} & 4.71 & 74.22 \\ [1ex]

        \qquad Friends-QIA & 60.06 \stdev{1.04} & 56.66 \stdev{1.22} & 3.40 & 46.61 \\
        \qquad IndirectQA EN + Friends-QIA  & 61.89 \stdev{1.61} & 55.67 \stdev{1.70} & 6.22 & 45.40 \\
        \qquad \geniqa\ EN \textsc{small} & 63.31 \stdev{8.67} & 57.72 \stdev{5.69} & 5.58 & 35.45 \\ [2ex]

        \textit{mDeBERTa trained with} & &  & \\
        \qquad IndirectQA EN & 48.08 \stdev{5.77} & 40.11 \stdev{2.86} & 7.97 & 35.61 \\
        \qquad \geniqa\ EN \textsc{small} & 68.13 \stdev{8.73} & 58.81 \stdev{4.62} & 9.32 & 35.45 \\
        \qquad \geniqa\ EN  & 65.11 \stdev{3.68} & 61.67 \stdev{3.61} & 3.44 & 31.80 \\
        \qquad \geniqa\ DE & 66.64 \stdev{1.08} & 56.56 \stdev{2.34} & 10.09 & 30.93 \\
        \qquad \geniqa\ BAR  & 65.58 \stdev{2.72} & 55.33 \stdev{1.00} & 10.24 & 31.20 \\ [2ex]

        \textit{XLM-R trained with} & &  & \\
        \qquad IndirectQA EN & 49.97 \stdev{12.80} & 45.53 \stdev{6.35} & 4.44 & 35.61 \\
        \qquad \geniqa\ EN \textsc{small} & 72.30 \stdev{5.84} & 59.89 \stdev{4.01} & 12.41 & 35.45 \\
        \qquad \geniqa\ EN  & 70.31 \stdev{0.92} & 62.44 \stdev{3.34} & 7.87 & 31.80 \\
        \qquad \geniqa\ DE  & 59.22 \stdev{5.16} & 53.67 \stdev{2.65} & 5.56 & 30.93 \\
        \qquad \geniqa\ BAR  & 67.16 \stdev{5.95} & 56.11 \stdev{4.11} & 11.04 & 31.20 \\
        
    \bottomrule
    \end{tabular}%
    }
    \caption{Average accuracy scores over three seeds on development and train datasets. The gap denotes the generalisation difference from the train accuracy to the dev accuracy.}
    \label{tab:train_dev_gap}
\end{table*}

To get the best possible performance for each model and training dataset, we train individual hyperparameter sets of learning rate, batch size warm-up ratio, weight decay, dropout rate and label smoothing for each setup (cf. Table \ref{tab:hyperparams} and \ref{tab:hyperparams_random}) as mentioned in \S\ref{sec:experimental_setup}. For this, we train the parameters in an order of importance to and influence on the model. This prioritization is necessary due to our available computational resources that do not allow testing all combinations. Thus, we divided the parameters into two fine-tuning groups: optimisation (learning rate, batch size and warm-up ratio) and regularisation (weight decay, dropout rate and label smoothing) parameters. We extract the best configuration of optimisation parameters depending on the accuracy score and use them to fine-tune the regularisation parameters to maximise the generalisation capabilities of the model in training. The final choice of parameters was again made depending on the best average accuracy score. Through the regularisation fine-tuning, we boost the accuracy on the dev data by an average of 0.5\,\% as opposed to only training the optimisation parameters.

Grid-searching is heavy on our computational resources. Finetuning mBERT \cite{devlin2019bert} takes $\sim$5 hours for small datasets like IndirectQA and $\sim$10 hours for large datasets like \geniqa, while fine-tuning mDeBERTa \cite{he2020mdeberta} and XLM-R \cite{conneau2020xmlroberta} take $\sim$18 hours for IndirectQA and $\sim$24 hours for \geniqa. All training runs and experiments were conducted with a Windows-based system with an 8GB VRAM Nvidia GeForce RTX 3070 GPU with and a Linux-based system with an Nvidia GeForce RTX 2060 SUPER with 8GB VRAM. 

Random hyperparameter search saves resources due to more efficient exploration of the hyperparameter space \cite{ulmer2022experimentalstandards}, which only takes $\sim$0.5 hours with IndirectQA and 30 trials, but the resulting parameters yielded slightly worse results with a deficit of -3 pps. for accuracy. We use random search for our experimental variants due to the training efficiency.

Each model has its own set of fine-tuned hyperparameters for the best performance because the parameters do not generalise well to other models with different architectures and parameter sizes. We observed this in pre-experiments where we trained models on IndirectQA EN with parameters fine-tuned on \geniqa\ BAR which resulted in lower accuracy scores for every evaluation dataset.

\section{Training and Development Dataset Results}\label{app:train_dev_results}

We present the results of our setups evaluated on their respective train and development datasets mentioned in \S\ref{subsec:train_and_dev_results} in Table \ref{tab:train_dev_gap}. The comparison of the accuracy scores and the gap between them on the train and dev sets reveals the overfitting. 
Usually, this generalisation error should be small \cite{zhang2016understanding}. However, for our trained models, we see that there is a large gap between each score-pair (cf. Table \ref{tab:train_dev_gap} with an average accuracy drop of -6 pps. from train to dev on mBERT \cite{devlin2019bert} models and -8 pps. on mDeBERTa \cite{he2020mdeberta} and XLM-R \cite{conneau2020xmlroberta}). 
The larger the gap between train and development results, the more overfitting took place.

All train and dev scores lie above the majority class baseline, but are low overall, even for English. We carefully inspected the training logs and confirm that the training loss goes down steadily. Due to overfitting, the results are not as expressive as we expected and need great care when analysing. 
High accuracy scores can be deceiving, as they often stem from an overprediction of the majority class and resemble the majority class baseline distribution. Further, high F1 scores are also not completely reliable. They can indicate that the model has learned good generalisation despite the low accuracy, but in some cases, we observed that the high F1 scores are caused by model biases for minority classes, which boosts the  macro-F1 through high recall scores. This is why we necessarily inspect both scores and check the predicted label distributions to inspect the specific model behaviour. 

High deviation scores for both accuracy and F1 scores (e.g., Tables \ref{tab:results_std}, \ref{tab:results_variants} and \ref{tab:train_dev_gap}), make the training and evaluation across various seeds a necessity. Otherwise, the performance would be too unreliable for analyses, as the scores vary
greatly.

\end{document}